# Variational Bayesian dropout: pitfalls and fixes

Jiri Hron [1]    Alexander G. de G. Matthews [1]    Zoubin Ghahramani [1 2]

## Abstract

Dropout, a stochastic regularisation technique for training of neural networks, has recently been reinterpreted as a specific type of approximate inference algorithm for Bayesian neural networks. The main contribution of the reinterpretation is in providing a theoretical framework useful for analysing and extending the algorithm. We show that the proposed framework suffers from several issues; from undefined or pathological behaviour of the true posterior related to use of improper priors, to an ill-defined variational objective due to singularity of the approximating distribution relative to the true posterior. Our analysis of the improper log uniform prior used in variational Gaussian dropout suggests the pathologies are generally irredeemable, and that the algorithm still works only because the variational formulation annuls some of the pathologies. To address the singularity issue, we proffer Quasi-KL (QKL) divergence, a new approximate inference objective for approximation of high-dimensional distributions. We show that motivations for variational Bernoulli dropout based on discretisation and noise have QKL as a limit. Properties of QKL are studied both theoretically and on a simple practical example which shows that the QKL-optimal approximation of a full rank Gaussian with a degenerate one naturally leads to the Principal Component Analysis solution.

## 1. Introduction

Srivastava et al. (2014) proposed dropout as a cheap way of preventing Neural Networks (NN) from overfitting. This work was rather impactful and sparked large interest in studying and extending the algorithm. One strand of this research lead to reinterpretation of dropout as a form of approximate Bayesian variational inference (Kingma et al., 2015; Gal & Ghahramani, 2016; Gal, 2016).

There are two main reasons for attempting reinterpretation of an existing method: 1) providing a principled interpretation of the empirical behaviour; 2) extending the method based on the acquired insights. Variational Bayesian dropout has been arguably successful in meeting the latter criterion (Kingma et al., 2015; Gal, 2016; Molchanov et al., 2017). This paper thus focuses on the former by studying the theoretical soundness of variational Bayesian dropout and the implications for interpretation of the empirical results.

The first main contribution of our work is identification of two main sources of issues in current variational Bayesian dropout theory:

(a) use of improper or pathological prior distributions;

(b) singularity of the approximate posterior distribution.

As we describe in Section 3, the log uniform prior introduced in (Kingma et al., 2015) generally does not induce a proper posterior, and thus the reported sparsification (Molchanov et al., 2017) cannot be explained by the standard Bayesian and the related minimum description length (MDL) arguments. In this sense, sparsification via variational inference with log uniform prior falls into the same category of non-Bayesian approaches as, for example, Lasso (Tibshirani, 1996). Specifically, the approximate uncertainty estimates do not have the usual interpretation, and the model may exhibit overfitting. Consequently, we study the objective from a non-Bayesian perspective, proving that the optimised objective is impervious to some of the described pathologies due to the properties of the variational formulation itself, which might explain why the algorithm can still provide good empirical results.[1]

Section 4 shows how mismatch between support of the approximate and the true posterior renders application of the standard Variational Inference (VI) impossible by making the Kullback-Leibler (KL) divergence undefined. As the second main contribution, we address this issue by proving that the remedies to this problem proposed in (Gal & Ghahramani, 2016; Gal, 2016) are special cases of a broader class of limiting constructions leading to a unique objective which we name Quasi-KL (QKL) divergence.

---

[1]An earlier version of this work was published in (Hron et al., 2017).

Section 5 provides initial discussion of QKL's properties, uses those to suggest an explanation for the empirically observed difficulty in tuning hyperparameters of the true model (e.g. Gal (2016, p. 119)), and demonstrates the potential of QKL on an illustrative example where we try to approximate a full rank Gaussian distribution with a degenerate one using QKL, only to arrive at the well known Principal Component Analysis (PCA) algorithm.

## 2. Background

Assume we have a discriminative probabilistic model $y \mid x, \boldsymbol{W} \sim \mathrm{P}(y \mid x, \boldsymbol{W})$ where $(x, y)$ is a single input-output pair, and $\boldsymbol{W}$ is the set of model parameters generated from a prior distribution $\mathrm{P}(\boldsymbol{W})$. In Bayesian inference, we usually observe a set of data points $(\boldsymbol{X}, \boldsymbol{Y}) = \{(x_n, y_n)\}_{n=1}^{\mathrm{N}}$ and aim to infer the posterior $p(\boldsymbol{W} \mid \boldsymbol{X}, \boldsymbol{Y}) \propto p(\boldsymbol{W}) \prod_n p(y_n \mid x_n, \boldsymbol{W})$,[2] which can be subsequently used to obtain the posterior predictive density $p(\boldsymbol{Y}' \mid \boldsymbol{X}', \boldsymbol{X}, \boldsymbol{Y}) = \int p(\boldsymbol{Y}' \mid \boldsymbol{X}', \boldsymbol{W}) p(\boldsymbol{W} \mid \boldsymbol{X}, \boldsymbol{Y}) \mathrm{d}\boldsymbol{W}$. If $p(y \mid x, \boldsymbol{W})$ is a complicated function of $\boldsymbol{W}$ like a neural network, both tasks often become computationally infeasible and thus we need to turn to approximations.

Variational inference approximates the posterior distribution over a set of latent variables $\boldsymbol{W}$ by maximising the evidence lower bound (ELBO),

$$\mathcal{L}(q) = \mathop{\mathbb{E}}_{\mathrm{Q}(\boldsymbol{W})}[\log p(\boldsymbol{Y} \mid \boldsymbol{X}, \boldsymbol{W})] - \mathrm{KL}\left(\mathrm{Q}(\boldsymbol{W}) \| \mathrm{P}(\boldsymbol{W})\right),$$

with respect to (w.r.t.) an approximate posterior $\mathrm{Q}(\boldsymbol{W})$. If $\mathrm{Q}(\boldsymbol{W})$ is parametrised by $\psi$ and the ELBO is differentiable w.r.t. $\psi$, VI turns inference into optimisation. We can then approximate the density of posterior predictive distribution using $q(\boldsymbol{Y}' \mid \boldsymbol{X}', \boldsymbol{X}, \boldsymbol{Y}) = \int p(\boldsymbol{Y}' \mid \boldsymbol{X}', \boldsymbol{W}) q(\boldsymbol{W}) \mathrm{d}\boldsymbol{W}$, usually by Monte Carlo integration.

A particular discriminative probabilistic model is a Bayesian neural network (BNN). BNN differs from a standard NN by assuming a prior over the weights $\boldsymbol{W}$. One of the main advantages of BNNs over standard NNs is that the posterior predictive distribution can be used to quantify uncertainty when predicting on previously unseen data $(\boldsymbol{X}', \boldsymbol{Y}')$. However, there are at least two challenges in doing so:

1) difficulty of reasoning about choice of the prior $\mathrm{P}(\boldsymbol{W})$;

2) intractability of posterior inference.

For a subset of architectures and priors, Item 1 can be addressed by studying limit behaviour of increasingly large networks (see, for example, (Neal, 1996; Matthews et al., 2018)); in other cases, sensibility of $\mathrm{P}(\boldsymbol{W})$ must be assessed individually. Item 2 necessitates approximate inference – a particular type of approximation related to dropout, the topic of this paper, is described below.

Dropout (Srivastava et al., 2014) was originally proposed as a regularisation technique for NNs. The idea is to multiply inputs of a particular layer by a random noise variable which should prevent co-adaptation of individual neurons and thus provide more robust predictions. This is equivalent to multiplying the rows of the subsequent weight matrix by the same random variable. The two proposed noise distributions were Bernoulli($p$) and Gaussian $\mathcal{N}(1, \alpha)$.

Bernoulli and Gaussian dropout were later respectively reinterpreted by Gal & Ghahramani (2016) and Kingma et al. (2015) as performing VI in a BNN. In both cases, the approximate posterior is chosen to factorise either over rows or individual entries of the weight matrices. The prior usually factorises in the same way, mostly to simplify calculation of KL$(\mathrm{Q}(\boldsymbol{W}) \| \mathrm{P}(\boldsymbol{W}))$. It is the choice of the prior and its interaction with the approximating posterior family that is studied in the rest of this paper.

## 3. Improper and pathological posteriors

Both Gal & Ghahramani (2016) and Kingma et al. (2015) propose using a prior distribution factorised over individual weights $w \in \boldsymbol{W}$. While the former opts for a zero mean Gaussian distribution, Kingma et al. (2015) choose to construct a prior for which KL$(\mathrm{Q}(\boldsymbol{W}) \| \mathrm{P}(\boldsymbol{W}))$ is independent of the mean parameters $\boldsymbol{\theta}$ of their approximate posterior $q(w) = \phi_{\theta, \alpha\theta^2}(w), w \in \boldsymbol{W}, \theta \in \boldsymbol{\theta}$, where $\phi_{\mu, \sigma^2}$ is the density function of $\mathcal{N}(\mu, \sigma^2)$. The decision to pursue such independence is motivated by the desire to obtain an algorithm that has no weight shrinkage – that is to say one where Gaussian dropout is the sole regularisation method. Indeed, the authors show that the log uniform prior $p(w) := \mathrm{C}/|w|$ is the only one where KL$(\mathrm{Q}(\boldsymbol{W}) \| \mathrm{P}(\boldsymbol{W}))$ has this mean parameter independence property. The log uniform prior is equivalent to a uniform prior on $\log|w|$. It is an improper prior (Kingma et al., 2015, p. 12) which means that there is no $\mathrm{C} \in \mathbb{R}$ for which $p(w)$ is a valid probability density.

Improper priors can sometimes lead to proper posteriors (e.g. normal Jeffreys prior for Gaussian likelihood with unknown mean and variance parameters) if $\mathrm{C}$ is treated as a positive finite constant and the usual formula for computation of posterior density is applied. We show this is generally not the case for the log uniform prior, and that any remedies in the form of proper priors that are in some sense close to the log uniform (such as uniform priors over floating point numbers) will lead to severely pathological inferences.

---

[2]Throughout the paper, $\mathrm{P}(\boldsymbol{W})$ refers to the distribution and $p(\boldsymbol{W})$ to its density function. Analogously for other distributions.

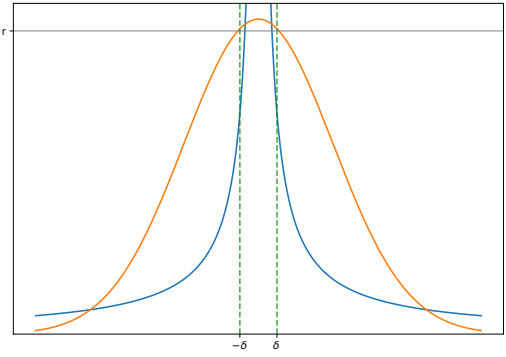
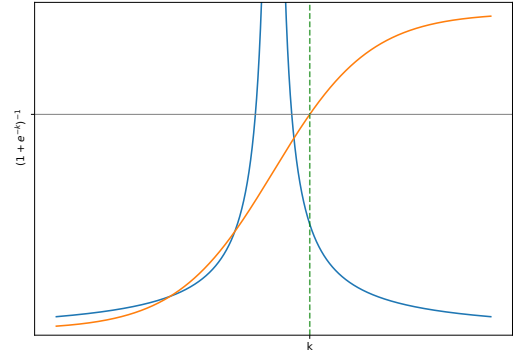

*Figure 1.* Illustration of Proposition 1. Blue is the prior, orange the likelihood, and green shows a particular neighbourhood of $w = 0$ where the likelihood is greater than $r > 0$ (such neighbourhood exists by the continuity). Integral of the likelihood over $(-\delta, \delta)$ w.r.t. $P(w)$ diverges because the likelihood can be lower bounded by $r > 0$ and the prior assigns infinite mass to this neighbourhood.

*Figure 2.* Visualisation of the infinite tail mass example. Blue is the prior, orange the sigmoid likelihood, and green shows the lower bound of the $[k, \infty)$ interval. The sigmoid function is greater than zero for any $k > 0$. The integral of the likelihood over $[k, \infty)$ w.r.t. $P(w)$ can thus again be lower bounded by a diverging integral.

### 3.1. Pathologies of the log uniform prior

For any proper posterior density, the normaliser $Z = \int_{\mathbb{R}^D} p(\boldsymbol{Y} \mid \boldsymbol{X}, \boldsymbol{W}) p(\boldsymbol{W}) d\boldsymbol{W}$ has to be finite (D denotes the total number of weights). We will now show that this requirement is generally not satisfied for the log uniform prior combined with commonly used likelihood functions.

**Proposition 1.** *Assume the log uniform prior is used and that there exists some $w \in \boldsymbol{W}$ such that the likelihood function at $w = 0$ is continuous in $w$ and non-zero. Then the posterior is improper.*

All proofs can be found in the appendix. Notice that standard architectures with activations like rectified linear or sigmoid, and Gaussian or Categorical likelihood satisfy the above assumptions, and thus the posterior distribution for non-degenerate datasets will generally be improper. See Figure 1 for a visualisation of this case.

Furthermore, the pathologies are not limited to the region near $w = 0$, but can also arise in the tails (Figure 2). As an example, we will consider a single variable Bayesian logistic regression problem $p(y \mid x, w) = 1/(1 + \exp(-xw))$, and again use the log uniform prior for $w$. For simplicity, assume that we have observed $(x = 1, y = 1)$ and wish to infer the posterior distribution. To show that the right tail has infinite mass, we integrate over $[k, \infty), k > 0$,

$$\int_{[k,\infty)} p(w) p(y \mid x, w) dw = \int_{[k,\infty)} \frac{C}{|w|} \frac{1}{1 + \exp(-w)} dw$$
$$> \int_{[k,\infty)} \frac{C}{|w|} \frac{1}{1 + \exp(-k)} dw = \frac{C \cdot (\infty - \log k)}{1 + \exp(-k)} = \infty.$$

Equivalently, we could have obtained infinite mass in the left tail, for example by taking the observation to be $(x = -1, y = 1)$. Because the sigmoid function is continuous and equal to $1/2$ at $w = 0$, the posterior also has infinite mass around the origin, exemplifying both of the discussed degeneracies. The normalising constant is of course still infinite and thus the posterior is again improper.

The practical implication of these pathologies is that even tasks as simple as MAP estimation (Proposition 1 implies unbounded posterior density) or posterior mean estimation will fail as the target is undefined. In general, improper posteriors lead to undefined or incoherent inferences. The above shows that this is the case for the log uniform prior combined with BNNs and related models, making Bayesian inference, exact and approximate, ill-posed.

### 3.2. Pathologies of the truncated log uniform prior

Neklyudov et al. (2017) proposed to swap the log uniform prior on $(-\infty, \infty)$ for a distribution that is uniform on a sufficiently wide bounded interval in the $\log|w|$ space (will be referred to as *the log space* from now on), i.e. $p(\log|w|) = 1/(b-a) \mathbb{I}_{[a,b]}(w), a < b$ where $\mathbb{I}_A$ is the indicator function of the set $A$. This prior can be used in place of the log uniform if the induced posteriors in some sense converge to a well-defined limit for any dataset as $[a, b]$ gets wider. If this is not the case, choice of $[a, b]$ becomes a prior assumption and must be justified as such because different choices will lead to sometimes considerably different inferences. We now show that posteriors generally do not converge for the truncated log uniform prior and discuss some of the related pathologies of the induced exact posterior.

To illustrate the considerable effect the choice of $[a, b]$ might have, we return to the example of posterior inference in a logistic regression model $p(y \mid x, w) = 1/(1 + e^{-xw})$ after observing $(x = 1, y = 1)$, using the prior $p_n(w) =$ $\mathbb{I}_{I_n}(w) \, C_n/|w|$ where $I_n = [-e^{b_n}, -e^{a_n}] \cup [e^{a_n}, e^{b_n}]$ (i.e. the appropriate transformation of the closed interval $[a_n, b_n]$ from *the log space* – see Figure 3). We exemplify the sensitivity of the posterior distribution to the choice of the $(I_n)_{n \in \mathbb{N}}$ sequence by studying the limiting behaviour of the posterior mean and variance. Using the definition of $\mathbb{I}_{I_n}(w)$ and symmetry, the normaliser of the posterior is,

$$Z_n = \int_{-e^{b_n}}^{-e^{a_n}} \frac{1}{|w|} \frac{1}{1+e^{-w}} dw + \int_{e^{a_n}}^{e^{b_n}} \frac{1}{|w|} \frac{1}{1+e^{-w}} dw$$
$$= \int_{e^{a_n}}^{e^{b_n}} \frac{1}{|w|} \frac{1+e^w}{1+e^w} dw = b_n - a_n \,.$$

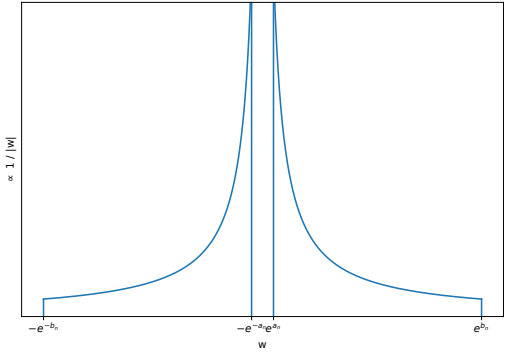

*Figure 3.* A truncated log uniform prior transformed to the original space. Notice that the support gap around the origin narrows as $a_n \to -\infty$, and the tail support expands as $b_n \to \infty$ which yields the more pathological inferences the wider $[a_n, b_n]$ gets.

Similar ideas can be used to derive the first two moments,

$$\mathbb{E}_{P_n}(w) = \frac{\int_{e^{a_n}}^{e^{b_n}} \frac{1}{1+e^{-w}} dw - \int_{-e^{b_n}}^{-e^{a_n}} \frac{1}{1+e^{-w}} dw}{b_n - a_n}$$
$$= \frac{h(e^{b_n}) + h(-e^{b_n}) - h(e^{a_n}) - h(-e^{a_n})}{b_n - a_n}, \quad (1)$$

$$\mathbb{E}_{P_n}(w^2) = \int_{e^{a_n}}^{e^{b_n}} \frac{|w|}{b_n - a_n} \frac{1+e^w}{1+e^w} dw = \frac{e^{2b_n} - e^{2a_n}}{2(b_n - a_n)}, \quad (2)$$

where $h(x) := \log(1+e^x)$, and $P_n$ stands for $P_n(w \mid x, y)$. To understand sensitivity of the posterior mean to the choice of $(I_n)_{n \in \mathbb{N}}$, we now construct sequences which respectively lead to convergence of the mean to zero, an arbitrary positive constant, and infinity.[3] To emphasise this is not specific to the posterior mean, we show that the variance might equally well be zero, infinite, or undefined.

To get $\lim_{n \to \infty} \mathbb{E}_{P_n}(w) = 0$, notice that for a fixed $b_n$, the second term in Equation (1) tends to $\log(4)/\infty = 0$.

---

[3]It would be equally possible to get convergence to an arbitrary negative constant, and negative infinity if the observation was $(x = -1, y = 1)$.

Hence we can make the posterior mean converge to zero by making the first term also tend to zero; a way to achieve this is setting $b_n = \log(\log|a_n|)$, which tends to infinity as $a_n \to \infty$. The limit of Equation (2) for the same sequence, and thus the variance, tends to zero as well.

For $\lim_{n \to \infty} \mathbb{E}_{P_n}(w) = c > 0$, we again focus on the first term in Equation (1) as the second term tends to zero for any increasing sequence $I_n \nearrow \mathbb{R}$. Simple algebra shows that for any diverging sequence $b_n \to \infty$, taking $a_n = b_n - e^{b_n}/c$ yields the desired result. The same sequence leads to infinite second moment and thus to infinite variance.

Finally, a choice which results in infinite mean and thus undefined variance is setting $a_n = -b_n$, for which the mean grows as $e^{b_n}/b_n$. We would like to point out that this symmetric growth of $a_n$ with $b_n$ is of particular interest as it corresponds to changing between different precisions of the float format representation on the computer as considered in Kingma et al. (2015, Appendix A).

### 3.3. Variational Gaussian dropout as penalised maximum likelihood

We have established that optimisation of the ELBO implied by a BNN with log uniform prior over its weights cannot generally be interpreted as a form of approximate Bayesian inference. Nevertheless, the reported empirical results suggest that the objective might possess reasonable properties. We thus investigate if and how the pathologies of the true posterior translate into the variational objective as used in (Kingma et al., 2015; Molchanov et al., 2017).

Firstly, we derive a new expression for $\mathrm{KL}(Q(w) \| P(w))$, and for its derivative w.r.t. the variational parameters, which will help us with further analysis.

**Proposition 2.** *Let $q(w) = \phi_{\mu, \sigma^2}(w)$, and $p(w) = C/|w|$. Denote $u := \mu^2/(2\sigma^2)$. Then,*

$$\mathrm{KL}(Q(w) \| P(w))$$
$$= \text{const.} + \frac{1}{2}\left(\log 2 + e^{-u} \sum_{k=0}^{\infty} \frac{u^k}{k!} \psi(1/2 + k)\right) \quad (3)$$

$$= \text{const.} - \frac{1}{2} \left.\frac{\partial M(a; 1/2; -u)}{\partial a}\right|_{a=0}, \quad (4)$$

*where $\psi(x)$ denotes the digamma function, and $M(a; b; z)$ the Kummer's function of the first kind.*

*We can obtain gradients w.r.t. $\mu$ and $\sigma^2$ using,*

$$\nabla_u \mathrm{KL}(Q(w) \| P(w)) = \begin{cases} 1 & u = 0 \\ \frac{D_+(\sqrt{u})}{\sqrt{u}} & u > 0 \end{cases}, \quad (5)$$

*and the chain rule; $D_+(x)$ is the Dawson integral. The derivative is continuous in $u$ on $[0, \infty)$.*

Before proceeding, we note that Equation (5) is sufficient to implement first order gradient-based optimisation, and thus can be used to replace the approximations used in (Kingma et al., 2015; Molchanov et al., 2017). Note that numerically accurate implementations of the $D_+(x)$ exist in many programming languages (e.g. (Johnson, 2012)).

In VI literature, the term $\mathrm{KL}\left(\mathrm{Q}(w) \| \mathrm{P}(w)\right)$ is often interpreted as a regulariser, constraining $\mathrm{Q}(w)$ from concentrating at the maximum likelihood estimate which would be optimal w.r.t. the other term $\mathbb{E}_{\mathrm{Q}(\boldsymbol{W})}[\log p(\boldsymbol{Y} \mid \boldsymbol{X}, \boldsymbol{W})]$ in the ELBO. It is thus natural to ask what effect this term has on the variational parameters. Noticing that only the infinite sum in Equation (3) depends on these parameters, and that the first summand is always equal to $\psi(1/2)$, we can focus on terms corresponding to $k \geq 1$. Because $\psi(1/2 + k) > 0, \forall k \geq 1$, all summands are non-negative. Hence the penalty will be minimised if $\mu^2/(2\sigma^2) = 0$, i.e. when $\mu = 0$ and/or $\sigma^2 \to \infty$; Corollary 3 is sufficient to establish that this minimum is unique.

**Corollary 3.** *Under assumptions of Proposition 2, $\mathrm{KL}\left(\mathrm{Q}(w) \| \mathrm{P}(w)\right)$ is strictly increasing for $u \in [0, \infty)$.*

Sections 3.1 and 3.2 suggests the pathological behaviour is non-trivial to remove unless we replace the (truncated) log uniform prior.[4] An alternative route is to interpret optimisation of the variational objective from above as a type of penalised maximum likelihood estimation.

Proposition 2 and Corollary 3 suggest that the variational formulation cancels the pathologies of the true posterior distribution which both invalidates the Bayesian interpretation, but also means that the algorithm may perform well in terms of accuracy and other metrics of interest. Since the $\mathrm{KL}\left(\mathrm{Q}(\boldsymbol{W}) \| \mathrm{P}(\boldsymbol{W})\right)$ regulariser will force the mean parameters to be small, and the variances to be large, and the $\mathbb{E}_{\mathrm{Q}(\boldsymbol{W})}[\log p(\boldsymbol{Y} \mid \boldsymbol{X}, \boldsymbol{W})]$ will generally push the parameters towards the maximum likelihood solution, the resulting fit might have desirable properties if the right balance between the two is struck. As the Bayesian interpretation no longer applies, the balance can be freely manipulated by reweighing the KL by any positive constant. The strict page limit and desire to discuss the singularity issue lead us to leave exploration of this direction to future work.

## 4. Approximating distribution singularities

Both the Bernoulli and Gaussian dropout can be seen as members of a larger family of algorithms where individual layer inputs are perturbed by elementwise i.i.d. random noise. This is equivalent to multiplying the corresponding row $\boldsymbol{w}_i$ of the subsequent weight matrix by the same noise variable. One could thus define $\boldsymbol{w}_i = s_i \boldsymbol{\theta}_i$, $s_i \sim \mathrm{Q}(s_i)$, $\mathrm{Q}(s_i)$ being an arbitrary distribution, and treat the induced distribution over $\boldsymbol{w}_i$ as an approximate posterior $\mathrm{Q}(\boldsymbol{w}_i)$.

---
[4]Louizos et al. (2017) made promising progress there.

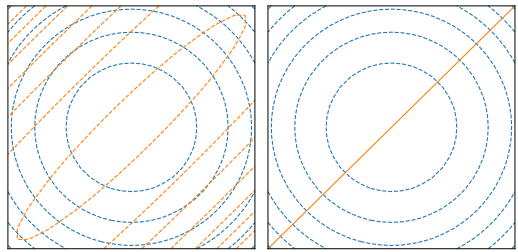

*Figure 4.* Illustration of approximating distribution singularities. On the left, blue is the standard and orange a correlated Gaussian density. *Null sets*, are (Borel) sets with zero measure under a distribution. Since both distributions have the same null sets, they are absolutely continuous w.r.t. each other. On the right, orange now represents a degenerate Gaussian supported on a line. Blue assigns zero probability to the line whereas orange assigns all of its mass; orange assigns probability zero to any set excluding the line but blue does not. Hence neither is absolutely continuous w.r.t. the other, and thus KL-divergence is undefined.

An issue with this approach is that it leads to undefined $\mathrm{KL}\left(\mathrm{Q}(\boldsymbol{W}) \| \mathrm{P}(\boldsymbol{W} \mid \boldsymbol{X}, \boldsymbol{Y})\right)$ whenever the prior assigns zero mass to the individual directions defined by $\boldsymbol{\theta}$. To understand why, note that $\mathrm{KL}\left(\mathrm{Q}(\boldsymbol{W}) \| \mathrm{P}(\boldsymbol{W} \mid \boldsymbol{X}, \boldsymbol{Y})\right)$ is defined only if $\mathrm{Q}(\boldsymbol{W})$ is *absolutely continuous* w.r.t. $\mathrm{P}(\boldsymbol{W} \mid \boldsymbol{X}, \boldsymbol{Y})$ which means that whenever $\mathrm{P}(\boldsymbol{W} \mid \boldsymbol{X}, \boldsymbol{Y})$ assigns probability zero to a particular set, $\mathrm{Q}(\boldsymbol{W})$ does so too. The right-hand side plot in Figure 4 shows a simple example of the case where neither distribution is absolutely continuous w.r.t. the other: the blue Gaussian assigns zero mass to any set with Lebesgue measure zero, such as the line along which the orange distribution places all its mass, and thus the orange Gaussian distribution is not absolutely continuous w.r.t. the blue one. This example is relevant to our problem from above, where $\mathrm{Q}(\boldsymbol{w}_i)$ always assigns all its mass to along the direction defined by the vector $\boldsymbol{\theta}_i$. For more details, see for example (Matthews, 2016, Section 2.1). When a measure is not absolutely continuous w.r.t. another measure, it can be shown to have a so called *singular* component relative to that measure, which we use as a shorthand for referring to this issue. Consequences for variational Bayesian interpretations of dropout are discussed next.

### 4.1. Implications for Bayesian dropout interpretations

Section 3.2 in (Kingma et al., 2015) proposes to use a shared Gaussian random variable for whole rows of the posterior weight matrices. Specifically $s_i \sim \mathcal{N}(1, \alpha)$ is substituted for $\mathrm{Q}(s_i)$ in the generic algorithm described in the previous section. We call such behaviour in the context of variational inference an *approximating distribution singularity*. The singularity has two possible negative consequences.

First, if only the $s_i$ scalars are treated as random variables, $\boldsymbol{\theta}$ become parameters of the discriminative model instead of the variational distribution. Optimisation of the ELBO will yield a valid Bayesian posterior approximation for the $s_i$. The lack of regularisation of $\boldsymbol{\theta}$ might lead to significant overfitting though, as $\boldsymbol{\theta}$ represent all weights in the BNN.

Second, if the fully factorised log uniform prior is used as before, then the directions defined by $\boldsymbol{\theta}$ constitute a measure zero subspace of $\mathbb{R}^D$, and thus the $\mathrm{KL}\left(\mathrm{Q}(\boldsymbol{W}) \| \mathrm{P}(\boldsymbol{W})\right)$ and consequently $\mathrm{KL}\left(\mathrm{Q}(\boldsymbol{W}) \| \mathrm{P}(\boldsymbol{W} \mid \boldsymbol{X}, \boldsymbol{Y})\right)$ are undefined for any configuration of $\boldsymbol{\theta}$. This is an instance of the issue described in the previous section. As a consequence, standard variational inference with this approximating family and target posterior is impossible.

A similar problem is encountered in (Gal & Ghahramani, 2016; Gal, 2016). The approximate posterior is defined as $\mathrm{Q}(\boldsymbol{w}_i) = p\, \delta_{\boldsymbol{0}} + (1-p)\, \delta_{\boldsymbol{\theta}_i}$ for each row in every weight matrix. The assumed prior is a product of independent non-degenerate Gaussian distributions which by definition assigns non-zero mass only to sets of positive Lebesgue measure. Again, the approximate posterior is not absolutely continuous w.r.t. the prior and thus the KL is undefined.

To address this issue, Gal & Ghahramani (2016) propose to replace the Dirac deltas in $\mathrm{Q}(\boldsymbol{w}_i)$ by Gaussian distributions with small but non-zero noise (we call this the *convolutional approach*). As an alternative, Gal (2016) proposes to instead discretise the Gaussian prior and the approximate posterior so both assign positive mass only to a shared finite set of values. Because the discretised Gaussian assigns non-zero mass to all points in the set, the approximate posterior is absolutely continuous w.r.t. this prior (we refer to this as the *discretisation approach*).

Strictly speaking, the two approaches cannot be equivalent because the corresponding random variables take values in distinct measurable spaces ($\mathbb{R}^D$ and a discrete grid respectively). Notwithstanding, both approaches are claimed to lead to the same optima for the variational parameters.[5] The suggested method for addressing this discrepancy is to introduce a *continuous relaxation* (Gal, 2016, p. 119) of the optimisation problem for the discrete case. The precise details of this relaxation are not given. One could define it as the relaxation that satisfied the required *KL-condition* (Gal, 2016, Appendix A), but there is of course then a risk of a circular argument. Putting these intuitive arguments on a firmer footing is one motivation for what follows here.

In the light of Section 3.2, it is natural to ask whether either of the proposed approaches will tend to a stable objective as the added noise shrinks to zero, and the discretisation becomes increasingly refined, respectively for the convolutional and discretisation approaches. Theorem 4 provides an affirmative answer by proving that both approaches lead to the same limit under reasonable assumptions.[6]

**Theorem 4.** *Let* $\mathrm{Q}, \mathrm{P}$ *be Borel probability measures on* $\mathbb{R}^D$, $\mathrm{P}$ *with a continuous density $p$ w.r.t. the D-dimensional Lebesgue measure, and $\mathrm{Q}$ supported on an at most countable measurable set* $S \subset \mathbb{Q}^D$, *with density $q$ w.r.t. the counting measure on* $\mathbb{Q}^D$. *If $S$ is infinite, further assume that* $\mathrm{diam}(S) < \infty$, *i.e.* $\sup_{x,y \in S} \|x-y\|_2 < \infty$.

*Then there exists a sequence* $(s^{(n)}) \subset \mathbb{R}$ *independent of* $\mathrm{Q}$ *and* $\mathrm{P}$ *s.t. the limit for both the sequences of convolved and discretised distributions* $\{(\mathrm{Q}^{(n)}, \mathrm{P}^{(n)})\}_{n \in \mathbb{N}}$,[7]

$$\lim_{n \to \infty} \left\{ \mathrm{KL}\left(\mathrm{Q}^{(n)} \| \mathrm{P}^{(n)}\right) - s^{(n)} \right\} = \mathbb{E}_{\mathrm{Q}}\left(\log \tfrac{q}{p}\right), \quad (6)$$

*given the perturbation noise is Gaussian and eventually shrinks to zero, and that the discretisation creates ever finer grid with equally sized cells as $n \to \infty$. The sequence* $(s^{(n)})$ *tends to 0 if* $\mathrm{Q} \ll \mathrm{P}$ *and to infinity otherwise.*

The right-hand side (r.h.s.) of Equation (6) satisfies Gal's *KL condition*, i.e. it leads to the same optimisation problem and thus unifies the convolutional and discretisation approach.

Unlike in (Gal, 2016, Appendix A), our derivation does not make an extraneous assumption on the distribution over any function of the $\boldsymbol{\theta}$ parameters nor does it require that the expectation of $\|\boldsymbol{\theta}_i\|_2^2$ grows without bounds with $\dim(\boldsymbol{\theta}_i)$. Neither of these two assumptions is sure to hold in practice as $\boldsymbol{\theta}$ are being optimised, and $\boldsymbol{\theta}_i$ in any modern (B)NN is initially scaled by $\sqrt{\dim(S)}$ exactly to achieve approximately constant Euclidean norm irrespective of the dimension.

We explored whether Equation (6) holds more generally. Theorem 5 extends the convolutional approach to a considerably larger class of approximating distributions.

**Theorem 5.** *Let* $\mathrm{Q}, \mathrm{P}$ *be Borel probability measures on* $\mathbb{R}^D$, $\mathrm{P}$ *with a bounded continuous density $p$ w.r.t. the Lebesgue measure on* $\mathbb{R}^D$, *and $\mathrm{Q}$ supported on a measurable linear manifold* $S \subset \mathbb{R}^D$ *of (Hamel) dimension* $\mathrm{K}_S$. *Assume $\mathrm{Q}$ has a continuous bounded density $q$ w.r.t. the Lebesgue measure on $S$, where the continuity is w.r.t. the trace topology.*

*Then there exists a sequence* $(s^{(n)}) \subset \mathbb{R}$ *dependent only on* $\mathrm{K}_S$ *s.t. the following holds for the convolutional approach,*

$$\lim_{n \to \infty} \left\{ \mathrm{KL}\left(\mathrm{Q}^{(n)} \| \mathrm{P}^{(n)}\right) - s^{(n)}_{\mathrm{K}_S} \right\} = \mathbb{E}_{\mathrm{Q}}\left(\log \tfrac{q}{p}\right), \quad (7)$$

*given the perturbation noise is Gaussian and eventually shrinks to zero. The sequence* $(s^{(n)}_{\mathrm{K}_S})$ *tends to 0 if* $\mathrm{Q} \ll \mathrm{P}$ *and to infinity otherwise.*

---

[5] Modulo the Euclidean distance to a closest point in the finite set for the discretisation approach.

[6] We state only the most important assumptions in Theorems 4 and 5. **Please see the appendix for the full set of assumptions**.

[7] $\mathrm{P}^{(n)} = \mathrm{P}$, $\forall n \in \mathbb{N}$, in the convolutional case.

A result related to Theorem 5 for the discretisation approach can be derived under assumptions similar to Theorem 4 with one important difference: $(s^{(n)}_{K_S})$, if it exists, is affected not only by $K_S$, but also by the orientation of $S$ in $\mathbb{R}^D$. This is because the dominating Lebesgue measure is different for each affine subspace $S$ and thus, unlike in the countable support case, $q$ cannot be defined w.r.t. a single dominating measure. Implicit in Theorems 4 and 5 is that the same constant can be subtracted from $\mathrm{KL}(\mathrm{Q}^{(n)} \| \mathrm{P}^{(n)})$ for all distributions Q with the same type of support. Hence if we are optimising over a family of singular approximating distributions, the sequence $(s^{(n)})$ (resp. $(s^{(n)}_{K_S})$) does not need to change between updates to obtain the desired limit.

Before moving to Section 5 which discusses some of the merits of using Equations (6) and (7) as an objective for approximate Bayesian inference, let us make two comments.

First, taking the limit makes the decision about size of perturbation or coarseness of the discretisation unnecessary. The sequences used do not cause the same instability problems discussed in Section 3.2 because the true posterior is well-defined even in the limit, which we assume in saying that P is a probability measure. The main open question is thus whether optimisation of the r.h.s. of Equation (6) will yield a sensible approximation of this posterior.

Second, if there is a family of approximate posterior distributions $\mathcal{Q}$ parametrised by $\psi \in \Psi$, the equality,

$$\underset{\psi \in \Psi}{\mathrm{argmin}}\ \underset{\mathrm{Q}_\psi}{\mathbb{E}}\left(\log \tfrac{q_\psi}{p}\right) = \lim_{n \to \infty} \underset{\psi \in \Psi}{\mathrm{argmin}}\ \mathrm{KL}(\mathrm{Q}^{(n)}_\psi \| \mathrm{P}^{(n)}), \tag{8}$$

need not hold unless stricter conditions are assumed. Equation (8) is of interest in cases when $\mathrm{KL}(\mathrm{Q}^{(n)}_\psi \| \mathrm{P}^{(n)})$ has some desirable properties (e.g. good predictive performance) which we would like to preserve. However, this is not the case for variational Bernoulli dropout as the objective being used by Gal & Ghahramani (2016) is, in terms of gradients w.r.t. the variational parameters, identical to the limit.

Furthermore, we can view both the discretisation and convolutional approaches as mere alternative vehicles to derive the same quasi discrepancy measure (cf. Section 5). If this quasi discrepancy possesses favourable properties, the precise details of optima attained along the sequence might be less important. One benefit of this view is in avoiding arguments like the previously mentioned *continuous relaxation* (Gal, 2016, p. 119).

## 5. Quasi-KL divergence

The r.h.s. of Equations (6) and (7) is markedly similar to the formula for standard KL divergence. We now make this link explicit. If $\mathrm{Z}_{\mathrm{P}_S} := \int_S p\, \mathrm{d}m_S < \infty$, $m_S$ being either the counting or the Lebesgue measure dominating measure for $q$, we can the probability density $p_S := p/\mathrm{Z}_{\mathrm{P}_S}$, and denote the corresponding distribution on $(S, \mathcal{B}_S)$ by $\mathrm{P}_S$. We term Equation (9) the *Quasi-KL* (QKL) divergence,

$$\mathrm{QKL}(\mathrm{Q} \| \mathrm{P}) := \underset{\mathrm{Q}}{\mathbb{E}}\left(\log \tfrac{q}{p}\right) = \mathrm{KL}(\mathrm{Q} \| \mathrm{P}_S) - \log \mathrm{Z}_{\mathrm{P}_S}. \tag{9}$$

Taking Equation (9) as a loss function says that we would like to find such a Q for which the KL divergence between Q and $\mathrm{P}_S$ is as small as possible, while making sure that the corresponding set $S$ runs through high density regions of P, preventing Q from collapsing to subspaces where $p$ is easily approximated by $q$ but takes low values. Since $p$ is continuous (c.f. Theorem 4), values of $p$ roughly indicate how much mass P assigns to the region where $S$ is placed.

Standard KL divergence and QKL are equivalent when $\mathrm{Q} \ll \mathrm{P}$ and the two distributions have the same support. QKL is not a proper statistical divergence though, as it is lower bounded by $-\log \mathrm{Z}_{\mathrm{P}_S}$ instead of zero. The non-negativity could have been satisfied by defining QKL as $\mathrm{KL}(\mathrm{Q} \| \mathrm{P}_S)$, dropping the $\log \mathrm{Z}_{\mathrm{P}_S}$ term. However, this would mean losing the above discussed effect of forcing $S$ to lie in a relatively high density region of P, and also the motivation of being a limit of the two sequences considered in Theorem 4.

Nevertheless, QKL inherits some of the attractive properties of KL divergence: the density $p$ need only be known up to a constant, the reparameterisation trick (Kingma & Welling, 2014) and analogical approaches for discrete random variables (Maddison et al., 2017; Jang et al., 2017; Tucker et al., 2017) still apply, and stochastic optimisation and integral approximation techniques can be deployed if desired.

On a more cautionary note, we emphasise that $\mathbb{E}_{\mathrm{Q}}(\log \tfrac{p}{q})$ is upper bounded by $\log \mathrm{Z}_{\mathrm{P}_S}$ and not the log marginal likelihood as is the case for standard KL use in VI. Hence optimisation of this objective w.r.t. hyperparameters of P need not work very well, since the resulting estimates could be biased towards regions where the variational family performs best.[8] This might explain why prior hyperparameters usually have to be found by validation error based grid search (Gal, 2016, e.g. p. 119) instead of ELBO optimisation as is common in the sparse Gaussian Process literature (Titsias, 2009).

Whether and when is QKL an attractive alternative to the more computationally expensive but proper statistical discrepancy measures which are capable of handling singular distributions (e.g. Wasserstein distances) is beyond the scope of this paper. To provide basic intuition of whether QKL might be a sensible objective for inference, Section 5.1 focuses on a simple practical example that yields a well known algorithm as the optimal solution to QKL optimisation, and exemplifies some of the above discussed behaviour.

---

[8]A similar issue for KL was observed by Turner et al. (2010).

## 5.1. QKL and Principal Component Analysis

Proposition 6 is an application of Theorem 5:

**Proposition 6.** *Assume* $P = \mathcal{N}(\mathbf{0}, \mathbf{\Sigma})$, $\mathbf{\Sigma}$ *a (strictly) positive definite matrix of rank* D, *with a degenerate Gaussian* $Q = \mathcal{N}(\mathbf{0}, \mathbf{A}\mathbf{V}\mathbf{A}^T)$, *where* $\mathbf{A}$ *is a* D × K *matrix with orthonormal columns, and* $\mathbf{V}$ *is a* K × K *(strictly) positive definite diagonal matrix. Then,*

$$\mathrm{QKL}(Q\|P) = c - \frac{1}{2}\sum_{k=1}^{K}\log \mathbf{V}_{kk} + \frac{1}{2}\mathrm{Tr}\left(\mathbf{A}^T\mathbf{\Sigma}^{-1}\mathbf{A}\mathbf{V}\right)$$

*where* c *is constant w.r.t.* $\mathbf{A}, \mathbf{V}$. *The optimal solution* $\mathbf{A}, \mathbf{V}$ *is to set columns of* $\mathbf{A}$ *to the top* K *eigenvectors of* $\mathbf{\Sigma}$ *and the diagonal of* $\mathbf{V}$ *to the corresponding eigenvalues.*[9]

Proposition 6 shows that the QKL-optimal way to approximate a full rank Gaussian with a degenerate one is to perform PCA on the covariance matrix. The result is intuitively satisfying as PCA preserves the directions of highest variance; $S$ was thus indeed forced to align with the highest density regions under P as suggested in Section 5. See Figure 5 for a visualisation of this behaviour. Proposition 7 presents a variation of the result of Tipping & Bishop (1999), showing that Equation (8) can hold in practice.

**Proposition 7.** *Assume similar conditions as in Proposition 6, except* Q *will now be replaced with a series of distributions convolved with Gaussian noise:* $Q^{(n)} = \mathcal{N}(\mathbf{0}, \mathbf{A}^{(n)}\mathbf{V}^{(n)}(\mathbf{A}^{(n)})^T + \tau^{(n)}\mathbf{I})$. *Given* $\tau^{(n)} \downarrow 0$ *as* $n \to 0$ *and the obvious constraints on* $\mathbf{A}^{(n)}, \mathbf{V}^{(n)}$, *Equation* (8) *holds in the sense of shrinking Euclidean/Frobenius norm between* $\{\mathbf{A}^{(n)}, \mathbf{V}^{(n)}\}$ *and the PCA solution.*

It is necessary to mention that both the QKL from Proposition 6 and any of the yet unconverged KL divergences in Proposition 7 have $\binom{D}{K}$ local optima for any combination of the eigenvectors which might lead to potentially problematic behaviour of gradient based optimisation.

## 6. Conclusion

The original intent behind dropout was to provide a simple yet effective regulariser for neural networks. The main value of the subsequent reinterpretation as a form of approximate Bayesian VI thus arguably lies in providing a principled theoretical framework which can explain the empirical behaviour, and guide extensions to the method. We have shown the current theory behind variational Bayesian dropout to have issues stemming from two main sources: 1) use of improper or pathological priors; 2) singular approximating distributions relative to the true posterior.

---

[9]We have assumed both Gaussians are zero mean to simplify the notation. Analogical results holds in the more general case.

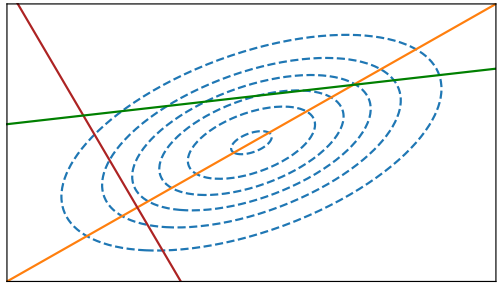

*Figure 5.* Visualisation of the relationship between QKL minimisation and PCA. The target in this example is the blue two dimensional Gaussian distribution. The approximating family is the set of all Gaussian distributions concentrated on a line, which would be problematic with conventional VI (c.f. Section 4). For all of the linear subspaces shown by the coloured lines the KL term on the right hand side of Equation (9) can be made zero by a suitable choice of the normal mean and variance. The remaining term $-\log Z_{P_S}$ therefore dictates the choice of subspace. The orange line is optimal aligning with the largest eigenvalue PCA solution.

The former issue pertains to the improper log uniform prior in variational Gaussian dropout. We proved its use leads to irremediably pathological behaviour of the true posterior, and consequently studied properties of the optimisation objective from a non-Bayesian perspective, arguing it is set up in such a way that cancels some of the pathologies and can thus still provide good empirical results, albeit not because of the Bayesian or the related MDL arguments.

The singular approximating distribution issue is relevant to both the Bernoulli and Gaussian dropout by making standard VI impossible due to an undefined objective. We have shown that the proposed remedies in (Gal & Ghahramani, 2016; Gal, 2016) can be made rigorous and are special cases of a broader class of limiting constructions leading to a unique objective which we termed quasi-KL divergence. We presented initial observations about QKL's properties, suggested an explanation for the empirical difficulty of obtaining hyperparameter estimates in dropout-based approximate inference, and motivated future exploration of QKL by showing it naturally yields PCA when approximating a full rank Gaussian with a degenerate one.

As use of improper priors and singular distributions is not isolated to the variational Bayesian dropout literature, we hope our work will contribute to avoiding similar pitfalls in future. Since it relaxes the standard KL assumptions, QKL will need further careful study in subsequent work. Nevertheless, based on our observations from Section 5 and the previously reported empirical results of variational Bayesian dropout, we believe QKL inspires a promising future research direction with potential to obtain a general framework for the design of computationally cheap optimisation-based approximate inference algorithms.

## Acknowledgements

We would like to thank Matej Balog, Diederik P. Kingma, Dmitry Molchanov, Mark Rowland, Richard E. Turner, and the anonymous reviewers for helpful conversations and valuable comments. Jiri Hron holds a Nokia CASE Studentship. Alexander Matthews and Zoubin Ghahramani acknowledge the support of EPSRC Grant EP/N014162/1 and EPSRC Grant EP/N510129/1 (The Alan Turing Institute).

# Appendix: Variational Bayesian dropout: pitfalls and fixes

Jiri Hron [1]   Alexander G. de G. Matthews [1]   Zoubin Ghahramani [1,2]

## A. Proofs for Section 3

*Notation and identities used throughout this section:* $\psi(x)$ for the digamma function, $\psi(x+1) = \psi(x) + 1/x$, $\psi(k+1) = H_k - \gamma$ where $H_k$ is the $k^{th}$ harmonic number and $\gamma$ is the Euler–Mascheroni's constant, $\text{Ei}(x) = -\int_{-x}^{\infty} e^{-t}/t \, dt$ is the exponential integral function, $\sum_{k=1}^{\infty} u^k H_k/k! = e^u(\gamma + \log u - \text{Ei}(-u))$ (Dattoli & Srivastava, 2008; Gosper, 1996), and $\sum_{k=1}^{\infty} u^k/(k!\,k) = \text{Ei}(u) - \gamma - \log u$ (Harris, 1957); the last two identities hold for $u > 0$. Importantly, we define $0^0 := 1$ unless stated otherwise.

*Proof of Proposition 1.* Denote the likelihood value by $\epsilon > 0$. Take an arbitrary number $r$ such that $\epsilon > r > 0$. By continuity, we can find $\delta > 0$ such that $|w - 0| < \delta$ implies that the likelihood value is greater than $r$; let $A \ni 0$ denote the open ball of radius $\delta$ centred at 0. Because both the prior density and the likelihood function only take non-negative values, we can apply the Tonelli–Fubini's theorem to obtain,

$$Z = \int_{\mathbb{R}^{D-1}} p(\boldsymbol{W}_{\neg w}) \left[ \int_{\mathbb{R}} p(w) p(\boldsymbol{Y} \mid \boldsymbol{X}, \boldsymbol{W}) \, dw \right] d\boldsymbol{W}_{\neg w}$$

$$> \int_{\mathbb{R}^{D-1}} p(\boldsymbol{W}_{\neg w}) \left[ \int_A \frac{C}{|w|} r \, dw \right] d\boldsymbol{W}_{\neg w} = \infty \,,$$

where $\boldsymbol{W}_{\neg w}$ is a shorthand for $\boldsymbol{W} \setminus w$. When $Z = \infty$, the measure of $\mathbb{R}^D$ under $P(\boldsymbol{W} \mid \boldsymbol{X}, \boldsymbol{Y})$ is infinite and thus it cannot be a proper probability distribution. □

*Proof of Proposition 2.* Using standard identities about Gaussian random variables, and the fact that $v := \varepsilon^2$, $\varepsilon \sim \mathcal{N}(\mu/\sigma, 1)$, follows the non-central chi-squared distribution $\chi^2(\lambda, \nu)$ with $\nu = 1$ degrees of freedom and non-centrality parameter $\lambda = (\mu/\sigma)^2$, we have,

$$\mathbb{E}_{Q(w)}[\log q(w)] - \mathbb{E}_{Q(w)}[\log p(w)]$$

$$= \mathbb{E}_{Q(w)}[\log q(w)] - \log C + \frac{1}{2} \mathbb{E}_{Q(w)}[\log |w|^2]$$

[1]Department of Engineering, University of Cambridge, Cambridge, United Kingdom [2]Uber AI Labs, San Francisco, California, USA. Correspondence to: Jiri Hron <jh2084@cam.ac.uk>.

$$= c_1 + \frac{1}{2} \mathbb{E}_{\varepsilon \sim \mathcal{N}(\mu/\sigma, 1)}[\log \sigma^2 \varepsilon^2]$$

$$= c_1 + \frac{1}{2} \left( \log \sigma^2 + \mathbb{E}_{v \sim \chi^2(\mu^2/\sigma^2, 1)}[\log v] \right)$$

$$= c_2 + \frac{1}{2} \int_0^\infty \sum_{k=0}^\infty e^{-\frac{\mu^2}{2\sigma^2}} \frac{(\frac{\mu^2}{2\sigma^2})^k}{k!} \frac{v^{k-\frac{1}{2}} e^{-\frac{v}{2}}}{2^{k+\frac{1}{2}} \Gamma(k+\frac{1}{2})} \log v \, dv \,,$$

where $c_1 := -\frac{1}{2} \log(2\pi e \sigma^2) - \log C$, $c_2 := c_1 + \frac{1}{2} \log(\sigma^2)$, and we used the fact that $\chi^2(\lambda, \nu)$ is equivalent to a Poisson mixture of centralised chi-squared distributions. Define,

$$f_n(v) := \sum_{k=0}^n e^{-\frac{\mu^2}{2\sigma^2}} \frac{(\frac{\mu^2}{2\sigma^2})^k}{k!} \frac{v^{k-\frac{1}{2}} e^{-\frac{v}{2}}}{2^{k+\frac{1}{2}} \Gamma(k+\frac{1}{2})} \log v \,,$$

and rewrite the last integral as,

$$\int_0^\infty \lim_{n \to \infty} f_n(v) dv$$
$$= \int_0^1 \lim_{n \to \infty} f_n(v) dv + \int_1^\infty \lim_{n \to \infty} f_n(v) dv \,.$$

Observe that $f_n \geq 0, \forall n \in \mathbb{N}$, and $f_n \uparrow f_\infty$ pointwise on $v \in [1, \infty)$, and $f_n < 0, \forall n \in \mathbb{N}$, and $f_n \downarrow f_\infty$ pointwise on $v \in [0, 1)$, for $f_\infty$ defined as the pointwise limit of $f_n$. Hence we can use the monotone convergence theorem as long as the $|\int f_0(v) dv| < \infty$. Using the identity $\mathbb{E}_{v \sim \chi^2(0, \nu)}[\log v] = \psi(\nu/2) - \log(1/2)$, we have,

$$\int_0^\infty f_n(v) dv = \log 2 + e^{-\frac{\mu^2}{2\sigma^2}} \sum_{k=0}^n \frac{(\frac{\mu^2}{2\sigma^2})^k}{k!} \psi(1/2 + k) \,,$$

which means that $f_n \in L^1, \forall n \in \mathbb{N}$. Because both $\int_0^1 |f_n(v)| dv$ and $\int_1^\infty |f_n(v)| dv$ are upper-bounded by $\int_0^\infty |f_n(v)| dv$, we can apply the monotone convergence theorem to equate,

$$\int_0^1 \lim_{n \to \infty} f_n(v) dv = \lim_{n \to \infty} \int_0^1 f_n(v) dv$$
$$\int_1^\infty \lim_{n \to \infty} f_n(v) dv = \lim_{n \to \infty} \int_1^\infty f_n(v) dv \,,$$

and thus by Theorem 4.1.10 in (Dudley, 2002) conclude $\int_0^\infty f_\infty(v) dv = \lim_{n \to \infty} \int_0^\infty f_n(v) dv$. Substituting back,

$$\mathbb{E}_{Q(w)}[\log q(w)] - \mathbb{E}_{Q(w)}[\log p(w)]$$

$$= c_2 + \frac{1}{2}\left(\log 2 + e^{-\frac{\mu^2}{2\sigma^2}} \sum_{k=0}^{\infty} \frac{(\frac{\mu^2}{2\sigma^2})^k}{k!} \psi(1/2 + k)\right)$$

$$= c_3 - \frac{1}{2} \left.\frac{\partial M(a; 1/2; -\mu^2/(2\sigma^2))}{\partial a}\right|_{a=0},$$

where $M(a; b; z)$ denotes the Kummer's function of the first kind, and $c_3 := c_2 - \frac{3}{2}\log 2 - \frac{1}{2}\gamma$. It is easy to check that Equation (3) holds for all $u = 0$ assuming $0^0 = 1$.

The last equality above was obtained using Wolfram Alpha (Wolfram—Alpha, 2017b). To validate this result, we performed an extensive numerical test, and will now show that the series indeed converges for $u = \mu^2/(2\sigma^2) \in [0, \infty)$, i.e. for all plausible values of $u$. The comparison test gives us convergence for $u \in (0, \infty)$:

$$\sum_{k=0}^{\infty} \frac{u^k}{k!} \psi(1/2 + k) < \psi(1/2) + \sum_{k=1}^{\infty} \frac{u^k}{k!} \psi(1 + k)$$

$$= \psi(1/2) + \sum_{k=1}^{\infty} \frac{u^k}{k!} (H_k - \gamma)$$

$$= \psi(1/2) + e^u(\gamma + \log u - \text{Ei}(-u)) - \gamma(e^u - 1)$$

$$= \psi(1/2) - \gamma + e^u(\log u - \text{Ei}(-u)),$$

where we use the fact that the individual summands are non-negative for $k \geq 1$ (which is also means we need not take the absolute value explicitly). It is trivial to check that the series converges at $u = 0$, and thus we have convergence for all $u \in [0, \infty)$.

To obtain the derivative with respect to $u$, we use the infinite series formulation from Equation (3), and the fact that the derivative of a power series within its radius of convergence is equal to the sum of its term-by-term derivatives (see (Gowers, 2014) for a nice proof). Using that only the infinite series in Equation (3) depends on $u$, we obtain,

$$\nabla_u e^{-u} \sum_{k=0}^{\infty} \frac{u^k}{k!} \psi(1/2 + k)$$

$$= \nabla_u \left(e^{-u}\psi(1/2) + e^{-u} \sum_{k=1}^{\infty} \frac{u^k}{k!} \psi(1/2 + k)\right)$$

$$= -e^{-u}\psi(1/2) + e^{-u} \sum_{k=1}^{\infty} \left(\frac{u^{k-1}}{(k-1)!}\psi(1/2 + k)\right)$$

$$\quad - e^{-u} \sum_{k=1}^{\infty} \left(\frac{u^k}{k!}\psi(1/2 + k)\right)$$

$$= e^{-u}(\psi(3/2) - \psi(1/2)) + e^{-u} \sum_{k=1}^{\infty} \left(\frac{u^k}{k!}\psi(3/2 + k)\right)$$

$$\quad - e^{-u} \sum_{k=1}^{\infty} \left(\frac{u^k}{k!}\psi(1/2 + k)\right)$$

$$= 2e^{-u} + e^{-u} \sum_{k=1}^{\infty} \frac{u^k}{k!} \frac{1}{1/2 + k} = e^{-u} \sum_{k=0}^{\infty} \frac{u^k}{k!} \frac{1}{1/2 + k}$$

$$= \frac{2D_+(\sqrt{u})}{\sqrt{u}},$$

for $u > 0$ and is equal to 2 if $u = 0$; in our case, the condition $u \geq 0$ is satisfied by definition; to obtain the expression in Equation (5), notice that the above series is multiplied by $1/2$ in Equation (3). Equality of the last infinite series to $2D_+(\sqrt{u})/\sqrt{u}$, was again obtained using Wolfram Alpha (Wolfram—Alpha, 2017a); the result was numerically validated, and convergence on $u \in (0, \infty)$ can again be established using the comparison test:

$$\sum_{k=0}^{\infty}\left|\frac{u^k}{k!}\frac{1}{1/2+k}\right| = \sum_{k=0}^{\infty}\frac{u^k}{k!}\frac{1}{1/2+k} < 2 + \sum_{k=1}^{\infty}\frac{u^k}{k!}\frac{1}{k}$$

$$= 2 + \text{Ei}(u) - \gamma - \log u.$$

The convergence at $u = 0$ is obtained trivially, yielding convergence for all $u \in [0, \infty)$.

$D_+(u)$ and $\sqrt{u}$ are continuous on $(0, \infty)$, and $\sqrt{u} > 0$; hence $D_+(u)/\sqrt{u}$ is continuous on $(0, \infty)$, and from definition of the Dawson integral $\lim_{u \to 0_+} D_+(\sqrt{u})/\sqrt{u} = 1$, i.e. the gradient is continuous in $u$ on $[0, \infty)$. □

*Proof of Corollary 3.* We use the conclusion of Proposition 2 which established differentiability for $u \in [0, \infty)$ (and thus continuity on the same interval). To show that $\text{KL}(\text{Q}(w)\|\text{P}(w))$ is strictly increasing for $u \in [0, \infty)$, it is sufficient to observe,

$$\nabla_u \text{KL}(\text{Q}(w)\|\text{P}(w)) = \frac{1}{2}e^{-u} \sum_{k=0}^{\infty} \frac{u^k}{k!} \frac{1}{1/2+k} > 0,$$

because each summand is strictly positive for $u \in [0, \infty)$ (given $0^0 = 1$). By a simple application of the mean value theorem, we conclude $\text{KL}(\text{Q}(w)\|\text{P}(w))$ is strictly increasing in $u$ on $[0, \infty)$. □

# B. Proofs for Section 4

Throughout this section, let $(\mathbb{R}^D, \|\cdot\|_2)$ be the D-dimensional Euclidean metric space, $\mathcal{T}$ the usual topology, and $\mathcal{B}$ the corresponding Borel $\sigma$-algebra. Let $\lambda^d$, $d \in \mathbb{N}$, be the d-dimensional Lebesgue measure.[1] P, Q will be probability measures, P with continuous density $p$ w.r.t. the Lebesgue measure on $\mathbb{R}^D$, and Q concentrated on some $S \in \mathcal{B}$, which is either (at most) countable or a linear manifold. Let $K_S$ be the Hausdorff dimension of $S$, i.e. zero in the countable, and $\dim(S)$ in the linear manifold case (dim being the Hamel dimension). The restriction $Q|_S$ of Q to $(S, \mathcal{B}_S)$, $\mathcal{B}_S$ the trace $\sigma$-algebra, will be denoted by $\widetilde{Q}$.

---

[1] More precisely the restriction of the m-dimensional Lebesgue measure to the corresponding Borel $\sigma$-algebra. We will be using the term Lebesgue measure instead of the sometimes used term *Borel measure* which we use to refer to any measure defined on the Borel $\sigma$-algebra.

Assume $\widetilde{Q}$ has a density $q$ w.r.t. the counting measure on $\mathbb{Q}^D$ if $S$ is at most countable,[2] or w.r.t. Lebesgue measure on $S$ in the linear manifold case. In the (at most) countable case, further assume that $\mathrm{diam}(S) < \infty$ if $S$ is infinite (trivially true if $S$ is finite). If $S$ is a linear manifold, assume that $q$ is continuous w.r.t. the trace topology $\mathcal{T}_S$, and that both $q$ and $p$ are bounded; denote the bounds on densities $q$ and $p$ by $C_q$ and $C_p$ respectively. We will be using $m_S$ as a shorthand for either of the corresponding dominating measures of $q$. We will also assume that $\log q \in \mathrm{L}^1(\widetilde{Q})$. Finally, the axiom of choice is assumed throughout.

We will be using the following fact: because $(\mathbb{R}^D, \|\cdot\|_2)$ is a complete separable metric space, every finite Borel measure is regular by Ulam's theorem (Dudley, 2002, Theorem 7.1.4), and thus tight by definition. Hence for any probability measure P on $(\mathbb{R}^D, \mathcal{B})$ and every $\varepsilon > 0$, there exists a compact set $C \in \mathcal{B}$ s.t. $P(C) > 1 - \varepsilon$.

The proofs of Theorems 4 and 5 will be divided into multiple propositions, each proven in a subsection corresponding to the limiting construction used.

*Proof of Theorem 4.* Combine Propositions 8 and 21. □

*Proof of Theorem 5.* Use Proposition 9. □

Notice that the statements of Propositions 8, 9 and 21 differ slightly from those of Theorems 4 and 5 by denoting the limit as $\mathbb{E}_{\widetilde{Q}} \log \frac{q}{p|_S}$ instead of $\mathbb{E}_Q \log \frac{q}{p}$. The former is more precise in the sense that $q$ is the density of $\widetilde{Q}$ w.r.t. $m_S$ on $(S, \mathcal{B}_S)$, and thus is not measurable w.r.t. Q, making the integral ill-defined. After swapping Q for $\widetilde{Q}$, the interchange of $p$ for $p|_S$ is necessary for similar reasons. We omitted this detail from the main text so as to meet the page limit, and to lighten the technicality of the discussion.

### B.1. Convolutional approach

Before approaching the proof of Propositions 8 and 9, we note that Lemma 11 allows us to assume that $S = \mathbb{R}^{K_S} \times \{0\}^{D-K_S}$ if $S$ is a linear manifold w.l.o.g.

The following definitions will be useful: let $Z$ and $\mathcal{E}$ be independent random variables respectively distributed according to the distributions $P_{\mathcal{E}} := \mathcal{N}(0, I_D)$ and Q. Define the shorthands $\mathcal{E}^{(n)} := \mathcal{E}/\sqrt{n}$ and $Z^{(n)} := Z + \mathcal{E}^{(n)}$. We further define the random variables $\widetilde{\mathcal{E}} := \mathcal{E}^{(n)}_{1:\,K_S} \times \{0\}^{D-K_S}$, where the subscript denotes the first $K_S$ elements of the vector ($\widetilde{\mathcal{E}}^{(n)} = 0$ if $K_S = 0$), $\widetilde{\mathcal{E}}^{(n)} := \widetilde{\mathcal{E}}/\sqrt{n}$, and $\widetilde{Z}^{(n)} := Z + \widetilde{\mathcal{E}}^{(n)}$. The corresponding distributions will be denoted as follows: $Q^{(n)} = \mathrm{Law}(Z^{(n)})$, $\widetilde{Q}^{(n)} := \mathrm{Law}(\widetilde{Z}^{(n)})$, $P_{\mathcal{E}}^{(n)} := \mathrm{Law}(\mathcal{E}^{(n)})$, $P_{\widetilde{\mathcal{E}}} := \mathrm{Law}(\widetilde{\mathcal{E}})$, and $P_{\widetilde{\mathcal{E}}}^{(n)} := \mathrm{Law}(\widetilde{\mathcal{E}}^{(n)})$.

Notice that $(Z, Z^{(n)}, \widetilde{Z}^{(n)})$ and $(\mathcal{E}^{(n)}, \widetilde{\mathcal{E}}^{(n)})$ are deterministically coupled collections of random variables. Also observe that we only convolve the approximating distribution with the Gaussian noise, and not the target P. Hence $P^{(n)} = P, \forall n \in \mathbb{N}$; we will thus omit the superscript here.

The convolution of two Borel measures $\mu, \nu$ on $\mathbb{R}^d$, $d \in \mathbb{N}$, will be denoted by $\mu \star \nu$ where for any measurable set $B$, $(\mu \star \nu)(B) = \int \mu(B - x)\nu(\mathrm{d}x)$. Observe $Q^{(n)} = Q \star \mathcal{N}_{\mathbb{R}^D}(0, n^{-1}I)$, and $\widetilde{Q}^{(n)} = \widetilde{Q} \star \mathcal{N}_S(0, n^{-1}I)$ with $\mathcal{N}_S(0, n^{-1}I) = P_{\widetilde{\mathcal{E}}}^{(n)}$ being the Gaussian probability measure on $(S, \mathcal{B}_S)$ (assuming $\mathcal{N}_S(\mu, \Sigma) = \delta_\mu$, the Dirac's delta distribution, if $S$ at most countable). As a corollary of (Dudley, 2002, Proposition 9.1.6), we have,

$$q^{(n)}(x) = \int \phi^{\lambda^D}_{x, n^{-1}I} q \, \mathrm{d}m_S \qquad , x \in \mathbb{R}^D, \qquad (10)$$

where $\phi^{\lambda^D}_{\mu, \Sigma}$ is the density function w.r.t. $\lambda^D$ of $\mathcal{N}(\mu, \Sigma)$ (we will omit the superscript unless confusion may arise). By an analogous argument, we obtain,

$$\widetilde{q}^{(n)}(x) = \int \phi^{m_S}_{x, n^{-1}I} q \, \mathrm{d}m_S \qquad , x \in S, \qquad (11)$$

where $\phi^{m_S}_{\mu, \Sigma}(z) = \delta_{\mathrm{Kr}}(z - \mu)$ if $S$ is at most countable ($\delta_{\mathrm{Kr}}$ is the Kronecker's delta function), as $m_S$ is the counting measure and $\mathcal{N}_S(\mu, \Sigma) = \delta_\mu$ (see above), and the usual density function of degenerate Gaussian if $m_S$ is the Lebesgue measure on $S$. Notice that it would have been more precise to replace $\phi^{\lambda^D}_{x, n^{-1}I}$ in Equation (10) with $\phi^{\lambda^D}_{x, n^{-1}I}|_S$ (c.f. Lemma 22); we omit the restriction in situations where its necessity is clear from the context.

**Proposition 8.** *Let $S$ be at most countable and all the relevant aforementioned assumptions hold. We consider two cases: $\log p \in \mathrm{L}^1(Q)$ and $\log p \notin \mathrm{L}^1(Q)$. If $\log p \in \mathrm{L}^1(Q)$, assume that the random variables $\{\log p(Z^{(n)})\}_{n \in \mathbb{N}}$ are uniformly integrable.*[3]

*Then,*

$$\lim_{n \to \infty} \left\{ \mathrm{KL}\left(Q^{(n)} \| P\right) - s^{(n)} \right\} = \mathbb{E}_{\widetilde{Q}} \log \frac{q}{p|_S},$$

*with $s^{(n)} := -\frac{D}{2} \log(2\pi e n^{-1})$.*

*Proof of Proposition 8.* First, assume that $\log p \in \mathrm{L}^1(Q)$. Because $\log q \in \mathrm{L}^1(\widetilde{Q})$ by assumption, we have $\log \frac{q}{p|_S} \in$ $L^1(\widetilde{Q})$ by Lemma 22 and (Dudley, 2002, Theorem 4.1.10). We can thus focus on convergence of the cross-entropy and negative entropy terms individually. By Lemma 12, the cross-entropy term converges. The negative entropy term converges by Lemma 13.

---

[2] We use the countable measure on rationals to avoid having to deal with a dominating measure that is not $\sigma$-finite.

[3] A useful sufficient condition is provided in Proposition 10.

It remains to investigate the case $\log p \notin L^1(Q)$. Because Lemma 13 still holds, we can invoke Lemma 20 which establishes that both the sequence $(KL(Q^{(n)} \| P) - s^{(n)})$ and the integral $\mathbb{E}_{\widetilde{Q}} \log \frac{q}{p|_S}$ do not converge as desired. $\square$

**Proposition 9.** *Let $S$ be a linear manifold and all the relevant aforementioned assumptions hold. We consider two cases: $\log p \in L^1(Q)$ and $\log p \notin L^1(Q)$. If $\log p \in L^1(Q)$, assume that the random variables $\{\log p(Z^{(n)})\}_{n \in \mathbb{N}}$ are uniformly integrable,[4] and that $\mathbb{E}\|Z\|_2^2 < \infty$.*

*Then,*

$$\lim_{n \to \infty} \left\{ KL(Q^{(n)} \| P) - s_{K_S}^{(n)} \right\} = \mathbb{E}_{\widetilde{Q}} \log \frac{q}{p|_S},$$

*with $s_{K_S}^{(n)} := -\frac{D - K_S}{2} \log(2\pi e n^{-1})$.*

*Proof of Proposition 9.* First, assume that $\log p \in L^1(Q)$. Because $\log q \in L^1(\widetilde{Q})$ by assumption, we have $\log \frac{q}{p|_S} \in L^1(\widetilde{Q})$ by Lemma 22 and (Dudley, 2002, Theorem 4.1.10). We can thus focus on convergence of the cross-entropy and negative entropy terms individually.

By Lemma 12, the cross-entropy term converges. Turning to the negative entropy term, by Lemma 14, we need to prove,

$$\mathbb{E} \log \widetilde{q}^{(n)}(\widetilde{Z}^{(n)}) \to \mathbb{E} \log q(Z).$$

Lemma 15 gives $\log \widetilde{q}^{(n)}(\widetilde{Z}^{(n)}) \to \log q(Z)$ a.s. Lemma 19 then yields the convergence in mean. Therefore,

$$\lim_{n \to \infty} \left\{ KL(Q^{(n)} \| P) - s_{K_S}^{(n)} \right\} = \mathbb{E}_{\widetilde{Q}} \log \frac{q}{p|_S}.$$

It remains to investigate the case $\log p \notin L^1(Q)$. Because Lemmas 14 and 15 and thus also Lemma 19 still hold, we can invoke Lemma 20 which establishes that both the sequence $(KL(Q^{(n)} \| P) - s^{(n)})$ and the integral $\mathbb{E}_{\widetilde{Q}} \log \frac{q}{p|_S}$ do not converge as desired. $\square$

**Proposition 10.** *For $f \in C(\mathbb{R}^D)$, a collection of random variables $\{f(Z^{(n)})\}_{n \in \mathbb{N}}$ is uniformly integrable if there exists some $r > 0$ s.t. $\forall x \in \mathbb{R}^D$ with $\|x\|_2 > r$, $|f(x)| \leq h_p(x)$ where $h_p \colon \mathbb{R}^D \to \mathbb{R}$, $x \mapsto \sum_{j=1}^p c_j \|x\|_2^j$, for some $c_1, \ldots, c_p \in \mathbb{R}$, and $\mathbb{E}\|Z\|_2^p < \infty$.[5]*

---

[4] A useful sufficient condition is provided in Proposition 10.

[5] Proposition 10 can be straightforwardly extended to polynomials in any $p$-norm $\|x\|_p = (\sum_{i=1}^D x_i^p)^{1/p}$, $p \in [1, \infty)$ by strong equivalence of $p$-norms on finite Euclidean spaces.

*Proof of Proposition 10.* Kallenberg (2006, p. 44, Equation (5)) states that a sequence of integrable random variables $\{\xi_n\}_{n \in \mathbb{N}}$ is uniformly integrable iff,

$$\lim_{k \to \infty} \limsup_{n \to \infty} \mathbb{E} \mathbb{I}_{|\xi_n| > k} |\xi_n| = 0. \quad (12)$$

Let us first ensure that random variables $\{f(Z^{(n)})\}_{n \in \mathbb{N}}$ are integrable. Defining $U := \{x \in \mathbb{R}^D \colon \|x\|_2 > r\}$,

$$\mathbb{E} \mathbb{I}_U |f(Z)| \leq \mathbb{E} \mathbb{I}_U h_p(Z),$$

with $h_p(Z)$ being a linear combination of terms $\|Z^{(n)}\|_2^k$ for $k \in 0, 1, \ldots, p$. By Cauchy–Bunyakovsky–Schwarz,

$$\mathbb{E} \mathbb{I}_U \|Z^{(n)}\|_2^k \leq \mathbb{E} \|Z + \mathcal{E}/\sqrt{n}\|_2^k$$
$$\leq 2^{\frac{3k}{2}-1} \left( \mathbb{E} \|Z\|_2^k + 2 \mathbb{E} \|Z\|_2^{\frac{k}{2}} \|\tfrac{\mathcal{E}}{\sqrt{n}}\|_2^{\frac{k}{2}} + \mathbb{E} \|\tfrac{\mathcal{E}}{\sqrt{n}}\|_2^k \right).$$

As $\mathbb{E}\|Z\|_2^t < \infty$ for all $t \in [0, p]$ by Hölder's inequality and the assumption $\mathbb{E}\|Z\|_2^p < \infty$, the second and third summands will go to 0 as $n \to \infty$, and the first term is finite. Because $\mathbb{E} \mathbb{I}_{U^C} |f(Z^{(n)})| \leq \sup_{U^C} |f|$ which is finite by continuity of $|f|$ and compactness of $U^C$ (Heine–Borel theorem), the random variables $\{f(Z^{(n)})\}_{n \in \mathbb{N}}$ are integrable.

By Equation (12), it is sufficient if $\forall \varepsilon > 0$, $\exists k \in \mathbb{R}$ s.t.,

$$\limsup_{n \to \infty} \mathbb{E} \mathbb{I}_{|f(Z^{(n)})| > k} |f(Z^{(n)})| < \varepsilon.$$

Because any finite collection of integrable random variables is uniformly integrable, we can find $\delta > 0$ s.t. $\forall B \in \mathcal{B}$ with $Q(B) \leq \delta$, $\mathbb{E} \mathbb{I}_B \|Z\|_2^j \leq \varepsilon/(2^{\frac{3j}{2}-1}|c_j|)$ for $j = 1, \ldots, p$. We w.l.o.g. assumed $c_j > 0, \forall j$ as otherwise we could just ignore the corresponding terms.

By tightness of Q, for every $\delta > 0$ there exists a compact set $K_{\delta,\alpha}$ s.t. $Q(K_{\delta,\alpha}) > 1 - \delta$ (the purpose of $\alpha$ will become clear later). Because we are on a finite Euclidean space, $K_{\delta,\alpha}$ is bounded and thus we can w.l.o.g. assume $K_{\delta,\alpha} = \bar{B}_{r_\delta - \alpha}(s_\delta)$, a closed ball centred at $s_\delta \in \mathbb{R}^D$ with radius $r_\delta - \alpha$, for some $\alpha > 0$, s.t. $r_\delta - \alpha > r$, i.e. $K_{\delta,\alpha}^C \subset U$. Clearly $K_{\delta,\alpha} \subset K_\delta := \bar{B}_{r_\delta}(s_\delta)$ and thus $Q(K_\delta) > 1 - \delta$. Define $\kappa = \sup_{K_\delta} |f|$ which is a finite constant by continuity of $f$ and compactness of $K_\delta$. We will now show,

$$\limsup_{n \to \infty} \mathbb{E} \mathbb{I}_{|f| > \kappa} |f(Z^{(n)})| < \varepsilon.$$

By the assumption $|f| \leq h_p$ on $U$, we have,

$$\mathbb{E} \mathbb{I}_{|f| > \kappa_\delta} |f(Z^{(n)})| \leq \mathbb{E} \mathbb{I}_{K_\delta^C} |f(Z^{(n)})|$$
$$\leq \sum_{j=1}^p c_j \, \mathbb{E} \mathbb{I}_{K_\delta^C} \|Z^{(n)}\|_2^j = \sum_{j=1}^p c_j \, \mathbb{E} \mathbb{I}_{K_\delta^C} \|Z + \mathcal{E}/\sqrt{n}\|_2^j,$$

where each of the r.h.s. summands can be upper bounded,

$$2^{\frac{3j}{2}-1} \left( \mathbb{E} \mathbb{I}_{K_\delta^C} \|Z\|_2^j + 2 \mathbb{E} \|Z\|_2^{\frac{j}{2}} \|\tfrac{\mathcal{E}}{\sqrt{n}}\|_2^{\frac{j}{2}} + \mathbb{E} \|\tfrac{\mathcal{E}}{\sqrt{n}}\|_2^j \right).$$

As before, all but the first term will vanish as $n \to \infty$ and thus we can ignore them in evaluation of the $\limsup$. Ignoring the multiplicative constants for a moment, we turn our attention to the $\mathbb{E}\,\mathbb{I}_{K_\delta^C}(Z^{(n)})\|Z\|_2^j = \mathbb{E}\,\mathbb{I}_{K_\delta^C}(Z + \mathcal{E}/\sqrt{n})\|Z\|_2^j$ where the noise term remained inside the indicator random variable by construction of the upper bound.

Define $A_\alpha^{(n)} \coloneqq \{x \in \mathbb{R}^D \colon \|x\|_2 \le \alpha\sqrt{n}\} \in \mathcal{B}$, $\beta^{(n)} \coloneqq P_\mathcal{E}(A_\alpha^{(n)})$ and observe $\beta^{(n)} \uparrow 1$. Because $\|Z + \mathcal{E}/\sqrt{n}\|_2 \le \|Z\|_2 + \|\mathcal{E}/\sqrt{n}\|_2$ by the triangle inequality, and $(Z + \mathcal{E}/\sqrt{n}) \in K_\delta^C$ iff $\|Z + \mathcal{E}/\sqrt{n}\|_2 > r_\delta$ by definition, we have $\mathbb{I}_{A_\alpha^{(n)}}(\mathcal{E})\,\mathbb{I}_{K_\delta^C}(Z + \mathcal{E}/\sqrt{n}) \le \mathbb{I}_{A_\alpha^{(n)}}(\mathcal{E})\,\mathbb{I}_{K_{\delta,\alpha}^C}(Z)$ for all $n \in \mathbb{N}$. Therefore,

$$\mathbb{E}[(\mathbb{I}_{A_\alpha^{(n)}}(\mathcal{E}) + \mathbb{I}_{(A_\alpha^{(n)})^C}(\mathcal{E}))\mathbb{I}_{K_\delta^C}(Z + \mathcal{E}/\sqrt{n})\|Z\|_2^j]$$
$$\le \mathbb{E}[\mathbb{I}_{A_\alpha^{(n)}}(\mathcal{E})\,\mathbb{I}_{K_{\delta,\alpha}^C}(Z)\|Z\|_2^j] + \mathbb{E}[\mathbb{I}_{(A_\alpha^{(n)})^C}(\mathcal{E})\|Z\|_2^j]$$
$$= \beta^{(n)}\,\mathbb{E}[\mathbb{I}_{K_{\delta,\alpha}^C}(Z)\|Z\|_2^j] + (1 - \beta^{(n)})\,\mathbb{E}\|Z\|_2^j.$$

Because $\mathbb{E}\|Z\|_2^j < \infty$ by Hölder's inequality and $\beta^{(n)} \uparrow 1$, the limit and thus $\limsup$ of the r.h.s. is clearly,

$$\mathbb{E}\,\mathbb{I}_{K_{\delta,\alpha}^C}(Z)\|Z\|_2^j < \frac{\varepsilon}{2^{\frac{3j}{2}-1}|c_j|},$$

where the upper bound is by uniform integrability of $\|Z\|_2^j$ and the construction of $K_{\delta,\alpha}$. Substituting back,

$$\limsup_{n \to \infty} \mathbb{E}\,\mathbb{I}_{|f| > k}|f(Z^{(n)})| < \varepsilon,$$

for all $k \ge \kappa$ which concludes the proof. $\square$

AUXILIARY LEMMAS

**Lemma 11.** *Assume $S$ is a linear manifold, i.e. $S = \{x \in \mathbb{R}^D \colon x = t(z), z \in S_0\}$, where $S_0 = \mathbb{R}^{K_S} \times \{0\}^{D-K_S}$, and $t \colon x \mapsto b + Ax$ with $b \in \mathbb{R}^D$ and $A \in \mathbb{R}^{D \times D}$ orthonormal. Then,*

$$\lim_{n \to \infty}\{\mathrm{KL}(Q^{(n)}\|P) - s^{(n)}\} = \mathbb{E}_{\widetilde{Q}} \log \frac{q}{p|_S},$$

*with $(s^{(n)}) \subset \mathbb{R}$, if and only if,*

$$\lim_{n \to \infty}\{\mathrm{KL}((t_\#^{-1}Q) \star P_\mathcal{E}^{(n)} \,\|\, t_\#^{-1}P) - s^{(n)}\}$$
$$= \mathbb{E}_{t_\#^{-1}\widetilde{Q}} \log \frac{q \circ t}{p|_S \circ t}.$$

*Furthermore,*

$$q \circ t = \frac{\mathrm{d}t_\#^{-1}\widetilde{Q}}{\mathrm{d}\lambda_{S_0}},$$

*where $\lambda_{S_0} = t_\#^{-1}m_S$ is the Lebesgue measure on $S_0$ with the corresponding trace $\sigma$-algebra $\mathcal{B}_{S_0}$. If $q$ is continuous and bounded, then also $q \circ t$ is continuous bounded w.r.t. the corresponding trace topology.*

*Proof of Lemma 11.* From definition, $t^{-1}(x) = A^\mathrm{T}(x - b)$ which is clearly a homeomorphism from $\mathbb{R}^D$ onto itself. Because we are working with Borel $\sigma$-algebras, we can use Lemma 7.5 in (Gray, 2011) to establish,

$$\mathrm{KL}(Q^{(n)}\|P) = \mathrm{KL}(t_\#^{-1}Q^{(n)}\|t_\#^{-1}P).$$

By definition, $t_\#^{-1}Q^{(n)} = \mathrm{Law}(t^{-1}(Z + \mathcal{E}^{(n)}))$; substituting $t^{-1}(Z + \mathcal{E}^{(n)}) = A^\mathrm{T}(Z + \mathcal{E}^{(n)} - b) = A^\mathrm{T}(Z - b) + A^\mathrm{T}\mathcal{E}^{(n)}$. Thus by properties of the multivariate normal distribution and orthonormality of $A$, $t_\#^{-1}Q^{(n)} = \mathrm{Law}(A^\mathrm{T}(Z - b) + A^\mathrm{T}\mathcal{E}^{(n)}) = \mathrm{Law}(A^\mathrm{T}(Z - b) + \mathcal{E}^{(n)}) = (t_\#^{-1}Q) \star P_\mathcal{E}^{(n)}$, and therefore for all $n \in \mathbb{N}$,

$$\mathrm{KL}(Q^{(n)}\|P) = \mathrm{KL}((t_\#^{-1}Q) \star P_\mathcal{E}^{(n)} \,\|\, t_\#^{-1}P).$$

Hence the two sequences of KL divergences are the same. By the substitution formula (see, for example, (Kallenberg, 2006, Lemma 1.22)),

$$\mathbb{E}_{t_\#^{-1}\widetilde{Q}} \log \frac{q \circ t}{p|_S \circ t} = \mathbb{E}_{\widetilde{Q}} \log \frac{q}{p|_S},$$

which finishes the first part of the proof.

Because $t$ is continuous, $q \circ t$ is continuous and bounded if the same holds for $q$. Finally, for any $B \in \mathcal{B}_{S_0}$,

$$t_\#^{-1}\widetilde{Q}(B) = \widetilde{Q}(t(B)) = \int_{t(B)} q\,\mathrm{d}m_S$$
$$= \int_S \mathbb{I}_B\left(t^{-1}(x)\right) q(x)\,t_\# \lambda_{S_0}(\mathrm{d}x)$$
$$= \int_S \mathbb{I}_B(x)\,q \circ t(x)\lambda_{S_0}(\mathrm{d}x) = \int_B q \circ t\,\mathrm{d}\lambda_{S_0},$$

which shows that $q \circ t = \frac{\mathrm{d}t_\#^{-1}\widetilde{Q}}{\mathrm{d}\lambda_{S_0}}$ as desired. $\square$

**Lemma 12.** *If $\{\log p(Z^{(n)})\}$ is uniformly integrable, then $\mathbb{E}_{Q^{(n)}} \log p \to \mathbb{E}_{\widetilde{Q}} \log p|_S$ as $n \to \infty$.*

*Proof of Lemma 12.* Notice that $\|Z^{(n)} - Z\|_2 = \|\mathcal{E}/\sqrt{n}\|_2$ by definition, and therefore $Z^{(n)} \to Z$ a.s. By the continuity of $p$ and of the logarithm function, the continuous mapping theorem yields $\log p(Z^{(n)}) \to \log p(Z)$ a.s. Since we have assumed that the collection of random variables $\{\log p(Z^{(n)})\}$ is uniformly integrable and a.s. convergence implies convergence in probability, we can use Theorem 10.3.6 in (Dudley, 2002) to deduce $\mathbb{E}_{Q^{(n)}} \log p \to \mathbb{E}_Q \log p$ as $n \to \infty$. By Lemma 22, $\mathbb{E}_Q \log p = \mathbb{E}_{\widetilde{Q}} \log p|_S$, concluding the proof. $\square$

**Lemma 13.** *If $S$ is at most countable, then,*

$$\lim_{n \to \infty}\{\mathbb{E}_{Q^{(n)}} \log q^{(n)} + \tfrac{D}{2}\log(2\pi e n^{-1})\} = \mathbb{E}_{\widetilde{Q}} \log q.$$

*Proof of Lemma 13.* The density of $\widetilde{Q}$ w.r.t. the counting measure on $\mathbb{Q}^D$ can be written using the Kronecker's delta function $\delta_{\mathrm{Kr}}$ as $q(x) = \sum_{i \in \mathbb{N}} \rho_i \delta_{\mathrm{Kr}}(x - m_i)$, where $\rho_i \geq 0$, $\sum_{i \in \mathbb{N}} \rho_i = 1$, and $m_i \in \mathbb{Q}^D$, $\forall i \in \mathbb{N}$. Recall that by Equation (10), the density of $Q^{(n)}$ w.r.t. $\lambda^D$ is,

$$q^{(n)}(x) = \sum_{i \in \mathbb{N}} \rho_i \, \phi_{m_i, n^{-1} I_D}(x) \,.$$

We can use the properties of multivariate normal distributions and the Tonelli–Fubini's theorem to establish,

$$\int q^{(n)} \log q^{(n)} \, d\lambda^D = -\frac{D}{2} \log(2\pi n^{-1}) +$$
$$\sum_{i \in \mathbb{N}} \int \rho_i \phi_{0, I_D}(\xi) \log \left[ \sum_{j \in \mathbb{N}} \rho_j e^{-\frac{\|m_i + \xi/\sqrt{n} - m_j\|_2^2}{2n^{-1}}} \right] \lambda^D(d\xi) \,,$$

which can be viewed as an integral over the product space $\mathbb{N} \times \mathbb{R}^D$ (remember $S$ is at most countable) w.r.t. the product measure of the distribution with density $i \mapsto \rho_i$ and the Gaussian $\mathcal{N}_{\mathbb{R}^D}(0, I)$. For any $i \in \mathbb{N}$ and $\xi \in \mathbb{R}^D$, define,

$$f^{(n)}(i, \xi) := \log \left[ \sum_{j \in \mathbb{N}} \rho_j \exp\left( -\frac{\|m_i + \xi/\sqrt{n} - m_j\|_2^2}{2n^{-1}} \right) \right].$$

Then $f^{(n)}(i, \xi) \to \log[\rho_i \exp(-\|\xi\|_2^2 / 2)] =: f^{(*)}(i, \xi)$ pointwise as $n \to \infty$. Furthermore, because the sum inside the logarithm is upper bounded by one, we have $|f^{(n)}(i, \xi)| = -f^{(n)}(i, \xi)$, $\forall n \in \mathbb{N}$, and since $-\log x \downarrow \infty$ as $x \downarrow 0$, we obtain $|f^{(n)}(i, \xi)| \leq -f^{(*)}(i, \xi)$ which is the negative logarithm of the $i^{\mathrm{th}}$ summand in $\exp[f^{(n)}(i, \xi)]$ for all $n \in \mathbb{N}$. Observing,

$$\sum_{i \in \mathbb{N}} \rho_i \mathop{\mathbb{E}}_{\xi \sim \mathcal{N}(0, I_D)} (f^{(*)}(i, \xi)) = -\frac{D}{2} + \sum_{i \in \mathbb{N}} \rho_i \log \rho_i \,,$$

we can invoke the dominated convergence theorem to establish (using the identity $-\frac{D}{2} = -\frac{D}{2} \log e$),

$$\int q^{(n)} \log q^{(n)} \, d\lambda^D + \frac{D}{2} \log(2\pi e n^{-1})$$
$$\to \sum_{i \in \mathbb{N}} \rho_i \log \rho_i = \mathop{\mathbb{E}}_{\widetilde{Q}} \log q \,,$$

as $n \to \infty$, concluding the proof. $\square$

**Lemma 14.** *For $S$ a linear manifold and every $n \in \mathbb{N}$, $\mathbb{E} \log q^{(n)}(Z^{(n)})$ is equal to,*

$$-\frac{D - K_S}{2} \log(2\pi e n^{-1}) + \mathbb{E} \log \widetilde{q}^{(n)}(\widetilde{Z}^{(n)}) \,.$$

*Proof of Lemma 14.* As stated at the beginning of this section, we can w.l.o.g. assume $S = \mathbb{R}^{K_S} \times \{0\}^{D - K_S}$. Then,

$$\log q^{(n)}(x)$$
$$= \log \left[ \int_{\mathbb{R}^D} (2\pi n^{-1})^{-\frac{D}{2}} e^{-\frac{\|x - z\|_2^2}{2n^{-1}}} Q(dz) \right]$$

$$= -\frac{D - K_S}{2} \log(2\pi n^{-1}) - \frac{n}{2} \left\| x_{(K_S+1):\,D} \right\|_2^2$$
$$+ \log \underbrace{\left[ \int_S \phi_{x_{1:K_S} \times \{0\}^{D-K_S}, n^{-1} I}^{m_S} \, d\widetilde{Q} \right]}_{= \widetilde{q}^{(n)}(x_{1:K_S} \times \{0\}^{D-K_S})},$$

$\forall x \in \mathbb{R}^D$, where we used Lemma 22 for the last equality. Using the definition $Z^{(n)} = Z + \mathcal{E}/\sqrt{n}$,

$$\mathbb{E} \log q^{(n)}(Z^{(n)})$$
$$= \iint \phi_{0, I}^{\lambda^D}(\xi) \log q^{(n)}(z + \xi/\sqrt{n}) \, \lambda^D(d\xi) Q(dz)$$
$$= -\frac{D - K_S}{2} \log(2\pi n^{-1}) - \frac{n}{2} \mathbb{E} \left\| \mathcal{E}_{(K_S+1):\,D}/\sqrt{n} \right\|_2^2$$
$$+ \iint \phi_{0, I}^{m_S}(\xi) \log \widetilde{q}^{(n)}(z + \xi/\sqrt{n}) \, m_S(d\xi) \widetilde{Q}(dz)$$
$$= -\frac{D - K_S}{2} \log(2\pi n^{-1}) - \frac{D - K_S}{2}$$
$$+ \iint \phi_{0, I}^{m_S}(\xi) \log \widetilde{q}^{(n)}(z + \xi/\sqrt{n}) \, m_S(d\xi) \widetilde{Q}(dz)$$
$$= -\frac{D - K_S}{2} \log(2\pi e n^{-1}) + \mathbb{E} \log \widetilde{q}^{(n)}(\widetilde{Z}^{(n)}) \,,$$

where the first equality is by the Tonelli–Fubini's theorem, the second by Lemma 22 and the standard marginalisation properties of the Gaussian distribution (the $\log \widetilde{q}^{(n)}$ term inside the integral only depends on $\mathcal{E}_{1:K_S}^{(n)}$), the third by the relation of independent Gaussian variables and the $\chi^2$ distribution, and the last again by the Tonelli-Fubini's theorem and the identity $-\frac{D - K_S}{2} = -\frac{D - K_S}{2} \log e$. $\square$

**Lemma 15.** *If $S$ is a linear manifold, then,*

$$\log \widetilde{q}^{(n)}(\widetilde{Z}^{(n)}) \to \log q(Z) \quad a.s.$$

*Proof.* Clearly $\widetilde{Z}^{(n)} = Z + \widetilde{\mathcal{E}}/\sqrt{n} \to Z$ a.s. Hence for fixed values $Z = z$ and $\widetilde{\mathcal{E}} = \xi$,

$$\left| \log \widetilde{q}^{(n)}(z + \xi/\sqrt{n}) - \log q(z) \right|$$
$$\leq \left| \log \widetilde{q}^{(n)}(z + \xi/\sqrt{n}) - \log q(z + \xi/\sqrt{n}) \right| \quad (13)$$
$$+ \left| \log q(z + \xi/\sqrt{n}) - \log q(z) \right|,$$

by the triangle inequality. The second term on the r.h.s. goes to zero with $n \to \infty$ by continuity of $q$. Turning to the first term, we can use the continuity of the logarithm to see that we only need to show that $\forall \varepsilon > 0$, $\exists N \in \mathbb{N}$ s.t. $|\widetilde{q}^{(n)}(z + \xi/\sqrt{n}) - q(z + \xi/\sqrt{n})| < \varepsilon$ for all $n \geq N$. Observe,

$$|\widetilde{q}^{(n)}(z + \tfrac{\xi}{\sqrt{n}}) - q(z + \tfrac{\xi}{\sqrt{n}})|$$
$$\leq \int \left| q(z + \tfrac{\xi + u}{\sqrt{n}}) - q(z + \tfrac{\xi}{\sqrt{n}}) \right| \mathcal{N}_S(0, I)(du) \,.$$

where $\mathcal{N}_S(\mu, \Sigma)$ is the Gaussian distribution on $S$ with the corresponding moments. Because $q$ is continuous, it is uniformly continuous on compact sets. Hence we can fix $\eta > 0$ and define $F := \bar{B}_{\|\xi\|_2 + \eta}(z)$, the closed ball centred at $z$ with radius $\|\xi\|_2 + \eta$, which is compact by the Heine–Borel theorem. Use uniform continuity to find $t > 0$ s.t. $\forall (x,y) \in F$ with $\|x - y\|_2 < t$ implies $|q(x) - q(y)| < \varepsilon$, and w.l.o.g. assume $t \le \eta$ (take $t = \eta$ if not). For $A := \{x \in S : \|x\|_2 < t\}$,

$$\int \left| q(z + \tfrac{\xi+u}{\sqrt{n}}) - q(z + \tfrac{\xi}{\sqrt{n}}) \right| \mathcal{N}_S(0, I)(\mathrm{d}u)$$
$$\le \int \mathbb{I}_A\left(\tfrac{u}{\sqrt{n}}\right) \left| q(z + \tfrac{\xi+u}{\sqrt{n}}) - q(z + \tfrac{\xi}{\sqrt{n}}) \right| \mathcal{N}_S(0, I)(\mathrm{d}u)$$
$$+ C_q \mathcal{N}_S(0, n^{-1}I)(A^C),$$

where the latter term on the r.h.s. vanishes as $n \to \infty$. Because $\|z + \tfrac{\xi+u}{\sqrt{n}} - z\|_2 \le \|\xi\|_2 + \|\tfrac{u}{\sqrt{n}}\|_2 < \|\xi\|_2 + t$ and $t \le \eta$, the first integral is clearly over a subset of $F$. Since $\|z + \tfrac{\xi+u}{\sqrt{n}} - z + \tfrac{\xi}{\sqrt{n}}\|_2 = \|\tfrac{u}{\sqrt{n}}\|_2$ which is lower than $t$ on $A$ by definition, the uniform continuity yields an upper bound,

$$|\widetilde{q}^{(n)}(z + \tfrac{\xi}{\sqrt{n}}) - q(z + \tfrac{\xi}{\sqrt{n}})| < \varepsilon + C_q \mathcal{N}_S(0, n^{-1}I)(A^C),$$

where the right hand side converges monotonically to $\varepsilon$ as desired. Therefore $\log \widetilde{q}^{(n)}(\widetilde{Z}^{(n)}) \to \log q(Z)$ a.s. $\square$

**Lemma 16.** *For $S$ a linear manifold and every $n \in \mathbb{N}$, $q^{(n)}$ and $\widetilde{q}^{(n)}$ are both bounded by the constant $C_q$ and continuous for $\mathcal{T}$ and $\mathcal{T}_S$ respectively.*

*Proof of Lemma 16.* Boundedness is a simple consequence of Equation (10) and the Hölder's inequality,

$$q^{(n)}(x) = \left\| \phi_{x, n^{-1}I}|_S \, q \right\|_{L^1(m_S)}$$
$$\le \left\| \phi_{x, n^{-1}I}|_S \right\|_{L^1(m_S)} \|q\|_{L^\infty(m_S)} = C_q;$$

similarly for $\widetilde{q}^{(n)}$ using Equation (11).

The proofs of continuity are analogous, therefore we will only discuss the one for $q$. Notice that for any $x, y \in \mathbb{R}^D$,

$$\left| q^{(n)}(x) - q^{(n)}(y) \right| \propto \left| \int f_z(x) - f_z(y) Q(\mathrm{d}z) \right|,$$

with $f_z(x) := \exp(-\tfrac{n}{2}\|x - z\|_2^2)$.

We can upper bound,

$$\left| \int f_z(x) - f_z(y) Q(\mathrm{d}z) \right| \le \int |f_z(x) - f_z(y)| \, Q(\mathrm{d}z),$$

which suggests it would be sufficient to show that the collection of functions $\{f_z\}_{z \in \mathbb{R}^D}$ is uniformly equicontinuous. A sufficient condition for uniform equicontinuity is $\{f_z\}_{z \in \mathbb{R}^D} \subset \mathrm{Lip}(\mathbb{R}^D, L)$ where $\mathrm{Lip}(\mathbb{R}^D, L)$ is the set of real-valued Lipschitz continuous functions on $\mathbb{R}^D$ with Lipschitz constant L. Because each $f_z$ is smooth, we can use Taylor expansion to equate,

$$f_z(x) = f_z(y) + (x - y)^T f'_z(\xi)$$

with $f'_z: \mathbb{R}^D \to \mathbb{R}^D$ the derivative of $f_z$, for some $\xi \in \mathbb{R}^D$. Using the Cauchy–Bunyakovsky–Schwarz inequality,

$$|f_z(x) - f_z(y)| \le \|x - y\|_2 \|f'_z(\xi)\|_2,$$

which means it is sufficient to show $\|f'_z(\xi)\|_2$ is uniformly bounded in $(z, \xi) \in \mathbb{R}^D \times \mathbb{R}^D$ to establish $\{f_z\}_{z \in \mathbb{R}^D} \subset \mathrm{Lip}(\mathbb{R}^D, L)$. Simple algebra shows that,

$$\|f'_z(\xi)\|_2 = n f_z(\xi) \|\xi - z\|_2 \le \sqrt{\tfrac{n}{e}},$$

$\forall (z, \xi) \in \mathbb{R}^D \times \mathbb{R}^D$, with equality when $\|\xi - z\|_2 = n^{-\frac{1}{2}}$. Hence we can see that $\{f_z\}_{z \in \mathbb{R}^D} \subset \mathrm{Lip}(\mathbb{R}^D, L)$ for $L = \sqrt{\tfrac{n}{e}}$, and thus the family of functions $\{f_z\}_{z \in \mathbb{R}^D}$ is uniformly equicontinuous.

Therefore, $\forall \varepsilon > 0, \exists \delta > 0$ s.t. $\|x - y\|_2 < \delta \implies |f_z(x) - f_z(y)| < \varepsilon$ for all $z \in \mathbb{R}^D$. Substituting back,

$$\left| q^{(n)}(x) - q^{(n)}(y) \right| < \left( \tfrac{n}{2\pi} \right)^{\tfrac{D}{2}} \varepsilon,$$

whenever $\|x - y\|_2 < \delta$, and thus $q^{(n)}$ is continuous. $\square$

**Lemma 17.** *For $S$ is a linear manifold, $\widetilde{q}^{(n)}$ converges pointwise to $q$ as $n \to \infty$.*

*Proof of Lemma 17.* W.l.o.g. assume $S = \mathbb{R}^{K_S} \times \{0\}^{D - K_S}$ (c.f. Lemma 11). For arbitrary $x \in S$,

$$\widetilde{q}^{(n)}(x) = \int q(x - \xi/\sqrt{n}) \mathcal{N}_S(0, I)(\mathrm{d}\xi),$$

where $\mathcal{N}_S(\mu, \Sigma)$ is the Gaussian measure on $S$ with the corresponding moments. Because $q$ is continuous by assumption, for every $\varepsilon > 0, \exists \delta > 0$ s.t. $\|(x - \xi/\sqrt{n}) - x\|_2 = \|\xi/\sqrt{n}\|_2 < \delta \implies |q(x - \xi/\sqrt{n}) - q(x)| < \varepsilon$. For any $\alpha > 0$, we can use Chebyshev's inequality to determine $N \in \mathbb{N}$ s.t. $\forall n \ge N, \mathbb{P}(\|\xi/\sqrt{n}\|_2 \ge \delta) \le \alpha$. Define $B \subset S$ to be the ball centred at zero with radius $\delta$. Observe,

$$\left| \widetilde{q}^{(n)}(x) - q(x) \right|$$
$$\le \int \left| q(x - \xi/\sqrt{n}) - q(x) \right| \mathcal{N}_S(0, I)(\mathrm{d}\xi)$$
$$< \varepsilon + \int_{B^C} \left| q(x - \xi/\sqrt{n}) - q(x) \right| \mathcal{N}(0, I_{K_S})(\mathrm{d}\xi)$$
$$\le \varepsilon + 2 C_q \alpha,$$

and therefore $\widetilde{q}^{(n)} \to q$ as $n \to \infty$ pointwise. $\square$

**Lemma 18.** *Assume $w_1, \ldots, w_k \in \mathbb{R}$ are arbitrary constants, and $\varepsilon_i$, $i = 1, \ldots, k$, are i.i.d. standard normal variables. Define the vector $w = (w_i)_{i=1}^k$. Then for $p \geq 0$,*

$$\mathbb{E}\left|\sum_{i=1}^k w_i \varepsilon_i\right|^p = \|w\|_2^p \frac{2^{\frac{p}{2}} \Gamma(\frac{p+1}{2})}{\Gamma(\frac{1}{2})}.$$

*Proof.* Use the linearity of the dot product and Gaussianity of $\varepsilon_i$'s to obtain,

$$\mathbb{E}\left|\sum_{i=1}^k w_i \varepsilon_i\right|^p = \mathbb{E}\left|\|w\|_2 \tilde{\varepsilon}\right|^p = \|w\|_2^p \, \mathbb{E}\, |\tilde{\varepsilon}|^p,$$

where $\tilde{\varepsilon}$ is a standard normal random variable. The result is then obtained by realising that powers of standard normal are distributed according to Generalised Gamma variable for which the expectation is known. $\square$

**Lemma 19.** *If $S$ is a linear manifold, $\mathbb{E}\|Z\|_2^2 < \infty$, and $\log \widetilde{q}^{(n)}(\widetilde{Z}^{(n)}) \to \log q(Z)$ a.s., then as $n \to \infty$,*

$$\mathbb{E} \log \widetilde{q}^{(n)}(\widetilde{Z}^{(n)}) \to \mathbb{E} \log q(Z).$$

*Proof of Lemma 19.* We define $Y := \log q(Z)$ and $\widetilde{Y}^{(n)} := \log \widetilde{q}^{(n)}(\widetilde{Z}^{(n)})$ and the corresponding probability measures $\nu := \mathrm{Law}(Y)$, $\nu^{(n)} := \mathrm{Law}(\widetilde{Y}^{(n)})$. Because a.s. convergence implies convergence in distribution, we have $\nu^{(n)} \to \nu$ weakly. Hence $\{\nu^{(n)}\}_{n \in \mathbb{N}}$ is uniformly tight by Proposition 9.3.4 in (Dudley, 2002), and so is $\{\nu^{(n)}\}_{n \in \mathbb{N}} \cup \{\nu\}$.

Therefore we can find a compact set $\bar{B}_\delta$ s.t. $\nu(\bar{B}_\delta) > 1 - \delta$ and $\nu^{(n)}(\bar{B}_\delta) > 1 - \delta, \forall n \in \mathbb{N}$ for any $\delta > 0$. W.l.o.g. we can assume that $\bar{B}_\delta$ is a closed interval as compactness is equivalent to closedness and boundedness for Euclidean spaces by the Heine–Borel theorem, and thus for any compact $\bar{B}_\delta$ we can find an interval $[s_\delta - r_\delta, s_\delta + r_\delta]$ satisfying the above condition for $\nu$ and all $\nu^{(n)}$.

Convergence in distribution implies that for any $f \in C_b(\mathbb{R})$, $\mathbb{E} f(\widetilde{Y}^{(n)}) \to \mathbb{E} f(Y)$ as $n \to \infty$. The identity function Id on $S$ is trivially continuous for the usual topology, but not bounded. However it is bounded on compact sets like $\bar{B}_\delta$. We thus approximate Id by a continuous compactly supported[6] function $h_{\delta,\eta} \mathrm{Id}$, for some fixed $\eta > 0$, where,

$$h_{\delta,\eta}(x) = \begin{cases} 1 & , \text{if } x \in \bar{B}_\delta \\ 0 & , \text{if } x \in F_{r,\eta} \\ \frac{r_\delta + \eta - |x - s_\delta|}{\eta} & , \text{else.} \end{cases}$$

with $F_{\delta,\eta}$ defined as complement of $(s_\delta - r_\delta - \eta, s_\delta + r_\delta + \eta)$.

Using the triangle inequality,

$$\left|\mathbb{E}_\nu(\mathrm{Id}) - \mathbb{E}_{\nu^{(n)}}(\mathrm{Id})\right| \leq \left|\mathbb{E}_\nu(\mathrm{Id}) - \mathbb{E}_\nu(h_{\delta,\eta}\mathrm{Id})\right|$$

---
[6]Support is the closure of the set where the function is non-zero.

$$+ \left|\mathbb{E}_\nu(h_{\delta,\eta}\mathrm{Id}) - \mathbb{E}_{\nu^{(n)}}(h_{\delta,\eta}\mathrm{Id})\right| + \left|\mathbb{E}_{\nu^{(n)}}(h_{\delta,\eta}\mathrm{Id}) - \mathbb{E}_{\nu^{(n)}}(\mathrm{Id})\right|.$$

Starting with the first term on the r.h.s., we can upper bound,

$$\left|\mathbb{E}_\nu(\mathrm{Id}) - \mathbb{E}_\nu(h_{\delta,\eta}\mathrm{Id})\right| \leq \mathbb{E}_\nu\left|(1 - h_{\delta,\eta})\mathrm{Id}\right| \leq \mathbb{E}_\nu \mathbb{I}_{\bar{B}_\delta^C}|\mathrm{Id}|,$$

and observe that $\mathbb{E}_\nu|\mathrm{Id}| \leq -\mathbb{E}_Q(\log \bar{q}) + |\log C_q|$, $\bar{q} := q/C_q$, which by $\log q \in L^1(Q)$ implies that $\mathrm{Id} \in L^1(\nu)$. Because any finite number of integrable functions is uniformly integrable, we can use Theorem 10.3.5 in (Dudley, 2002) to conclude that $\forall \varepsilon > 0$, there exists $\delta > 0$ s.t. $\mathbb{E}_\nu \mathbb{I}_{\bar{B}_\delta^C}|\mathrm{Id}| \leq \varepsilon$. Denote this number by $\delta_1$.

Turning to the last term, we can again upper bound $\left|\mathbb{E}_{\nu^{(n)}}(h_{\delta,\eta}\mathrm{Id}) - \mathbb{E}_{\nu^{(n)}}(\mathrm{Id})\right|$ with $\mathbb{E}_{\nu^{(n)}} \mathbb{I}_{\bar{B}_\delta^C}|\mathrm{Id}|$, $\forall n \in \mathbb{N}$. In this case, it will be beneficial to revert to the original representation:

$$\mathbb{E}_{\nu^{(n)}} \mathbb{I}_{\bar{B}_\delta^C}|\mathrm{Id}| = \mathbb{E}_{\widetilde{Q}^{(n)}} \mathbb{I}_{(A_\delta^{(n)})^C} |\log \widetilde{q}^{(n)}|,$$

with $A_\delta^{(n)} := (\log \widetilde{q}^{(n)})^{-1}(\bar{B}_\delta)$; observe that because $\nu^{(n)} = (\log \widetilde{q}^{(n)})_\# \widetilde{Q}^{(n)}$, $\widetilde{Q}^{(n)}(A_\delta^{(n)}) > 1 - \delta, \forall n \in \mathbb{N}$, by definition. By Lemma 16, each $\widetilde{q}^{(n)}$ is bounded by $C_q$, thus we w.l.o.g. assume that $|\log \widetilde{q}^{(n)}| = -\log \widetilde{q}^{(n)}$ as the normalisation by $C_q$ will only add a vanishing term $C_q \widetilde{Q}^{(n)}((A_\delta^{(n)})^C) \leq C_q \delta$ on the r.h.s., $\forall n \in \mathbb{N}$. Then,

$$\mathbb{E}_{\widetilde{Q}^{(n)}} \mathbb{I}_{(A_\delta^{(n)})^C} |\log \widetilde{q}^{(n)}|$$
$$= - \mathbb{E}_{\widetilde{Q}^{(n)}} (\mathbb{I}_{(A_\delta^{(n)})^C} \log \widetilde{q}^{(n)}) \pm \mathbb{E}_{\widetilde{Q}^{(n)}} (\mathbb{I}_{(A_\delta^{(n)})^C} \log \phi_{0,I}^{m_S})$$
$$= - \mathbb{E}_{\widetilde{Q}^{(n)}} \left(\mathbb{I}_{(A_\delta^{(n)})^C} \log \frac{\widetilde{q}^{(n)}}{\phi_{0,I}^{m_S}}\right) - \mathbb{E}_{\widetilde{Q}^{(n)}} (\mathbb{I}_{(A_\delta^{(n)})^C} \log \phi_{0,I}^{m_S})$$
$$\leq - \widetilde{Q}^{(n)}((A_\delta^{(n)})^C) \log \frac{\widetilde{Q}^{(n)}((A_\delta^{(n)})^C)}{\mathcal{N}_S(0,I)((A_\delta^{(n)})^C)}$$
$$\quad - \mathbb{E}_{\widetilde{Q}^{(n)}} (\mathbb{I}_{(A_\delta^{(n)})^C} \log \phi_{0,I}^{m_S}),$$

where the inequality is by Equation (7) on p. 177 in (Gray, 2011), and the fact that non-degenerate Gaussian distributions on Euclidean spaces are *equivalent* to the corresponding Lebesgue measure (i.e. $\mathcal{N}(\mu, \Sigma) \ll \lambda^k$ and $\lambda^k \ll \mathcal{N}(\mu, \Sigma)$ for all $k \in \mathbb{N}, \mu \in \mathbb{R}^k$ and positive definite $\Sigma$) which means that $\widetilde{Q}^{(n)} \ll \mathcal{N}_S(0, I), \forall n \in \mathbb{N}$, and thus,

$$\mathbb{E}_{\widetilde{Q}^{(n)}} \mathbb{I}_{(A_\delta^{(n)})^C} \log \frac{\widetilde{q}^{(n)}}{\phi_{0,I}^{m_S}}$$

in the above derivation is well-defined. $\widetilde{Q}^{(n)} \ll \mathcal{N}_S(0, I)$ implies that $\widetilde{Q}^{(n)}((A_\delta^{(n)})^C) > 0$ if $\mathcal{N}(0, I_S)((A_\delta^{(n)})^C) > 0$ meaning we can upper bound the first term on the r.h.s. by,

$$-\widetilde{Q}^{(n)}((A_\delta^{(n)})^C) \log \widetilde{Q}^{(n)}((A_\delta^{(n)})^C),$$

which vanishes as $\delta \to 0$. The second term is equal to,

$$-\widetilde{Q}^{(n)}((A_\delta^{(n)})^C)\tfrac{K_S}{2}\log(2\pi) - \tfrac{1}{2}\mathbb{E}\,\mathbb{I}_{(A_\delta^{(n)})^C}\left\|Z+\widetilde{\mathcal{E}}/\sqrt{n}\right\|_2^2,$$

where the first term again vanishes as $\delta \to 0$. Combining $\Gamma(0) = 1$, $\Gamma(\tfrac{1}{2}) = \sqrt{\pi}$ and Lemma 18, the latter term can be upper bounded by,

$$\mathbb{E}(\mathbb{I}_{(A_\delta^{(n)})^C}\|Z\|_2^2) + \frac{\mathbb{E}\|Z\|_2}{\sqrt{2\pi n}} + \frac{\mathbb{E}\|\widetilde{\mathcal{E}}\|_2^2}{n}.$$

As $\mathbb{E}\|\widetilde{\mathcal{E}}\|_2^2 = K_S$, the last term will vanish as $n \to \infty$. Because we have assumed $\mathbb{E}\|Z\|_2^2 < \infty$, Hölder's inequality yields $\mathbb{E}\|Z\|_2 < \infty$ and thus the second term will also disappear as $n \to \infty$. $\mathbb{E}\|Z\|_2^2 < \infty$ can also be used to determine that the singleton set $\{\|Z\|_2^2\}$ is uniformly integrable and thus again by Theorem 10.3.5 in (Dudley, 2002) $\mathbb{E}\,\mathbb{I}_{(A_\delta^{(n)})^C}\|Z\|_2^2 \to 0$ as $\delta \to 0$. Notice that the terms that vanish with $\delta \to 0$ do so independently of $n$ by uniform tightness of $\{\widetilde{Q}^{(n)}\}_{n\in\mathbb{N}}$ and the construction of $A_\delta^{(n)}$. We can thus find constants $N_1 \in \mathbb{N}$ and $\delta_2 > 0$ which will make $\mathbb{E}_{\nu^{(n)}}\mathbb{I}_{\bar{B}_\delta^C}|\mathrm{Id}|$, $n \geq N$, arbitrarily small.

Finally, the second term in our original upper bound, $|\mathbb{E}_\nu(h_{\delta,\eta}\mathrm{Id}) - \mathbb{E}_{\nu^{(n)}}(h_{\delta,\eta}\mathrm{Id})|$ will tend to zero as $n \to \infty$ for fixed $\delta > 0$ and $\eta > 0$ as $h_{\delta,\eta}\mathrm{Id} \in C_b(\mathbb{R})$. $\eta$ is only introduced for $h_{\delta,\eta}\mathrm{Id}$ to be a continuous compactly supported function and thus can be set to an arbitrary positive number. Setting $\delta = \delta_1 \wedge \delta_2$, we can thus find $N_2 \in \mathbb{N}$ that will make $|\mathbb{E}_\nu(h_{\delta,\eta}\mathrm{Id}) - \mathbb{E}_{\nu^{(n)}}(h_{\delta,\eta}\mathrm{Id})|$ arbitrarily small.

To establish that $|\mathbb{E}_\nu(\mathrm{Id}) - \mathbb{E}_{\nu^{(n)}}(\mathrm{Id})|$ can be made arbitrarily small, simply take $N = N_1 \vee N_2$. Hence $\mathbb{E}\log\widetilde{q}^{(n)}(\widetilde{Z}^{(n)}) \to \mathbb{E}\log q(Z)$ as $n \to \infty$. $\square$

**Lemma 20.** *If* $\log p \notin L^1(Q)$*, and,*

$$\lim_{n\to\infty}\left\{\mathbb{E}_{Q^{(n)}}\log q^{(n)} - s^{(n)}\right\} = \mathbb{E}_{\widetilde{Q}}\log q,$$

*then* $\mathbb{E}_{\widetilde{Q}}\log\frac{q}{p|_S}$ *and* $(\mathbb{E}_{Q^{(n)}}\log\frac{q}{p|_S} - s^{(n)})$ *diverge.*

*Proof of Lemma 20.* By Lemma 22, $\log p|_S \notin L^1(\widetilde{Q})$. Because $\log q \in L^1(\widetilde{Q})$ by assumption, we have $\log\frac{q}{p|_S} \notin L^1(\widetilde{Q})$ which yields the first part of the claim.

We now turn to the second part, i.e. to the sequence $(\mathbb{E}_{Q^{(n)}}\log\frac{q}{p|_S} - s^{(n)})$.

First, we prove that $\mathbb{E}_{Q^{(n)}}\log p$ cannot converge. Since $p$ is bounded, we can w.l.o.g. assume $\log p \leq 0$. If $\log p \notin L^1(Q^{(n)})$ infinitely often, $\mathbb{E}\log p(Z^{(n)})$ does not converge. Otherwise $\log p \in L^1(Q^{(n)})$, $\forall n \geq N$, for some $N \in \mathbb{N}$. Notice that $Z^{(n)} \to Z$ a.s., and by continuity of $\log p$, also $\log p(Z^{(n)}) \to \log p(Z)$ a.s. Because the zero function is trivially integrable and $\log p \leq 0$, we can use the reverse Fatou's lemma to establish,

$$\limsup_{n\to\infty}\mathbb{E}\log p(Z^{(n)}) \leq \mathbb{E}\log p(Z) = -\infty,$$

where we have used that $\log p \leq 0$ and $\log p \notin L^1(Q)$ in the last equality. Again $\mathbb{E}\log p(Z^{(n)})$ does not converge.

Now we need to prove $(\mathbb{E}_{Q^{(n)}}\log\frac{q^{(n)}}{p} - s^{(n)})$ does not converge. Assume the sequence converges to some $\kappa \in \mathbb{R}$. By assumption $(\mathbb{E}_{Q^{(n)}}\log q^{(n)} - s^{(n)})$ converges. Thus by (Dudley, 2002, Theorem 4.1.10), $\mathbb{E}_{Q^{(n)}}\log p$ must also converge which is a contradiction of the divergence established above. Therefore $(\mathbb{E}_{Q^{(n)}}\log\frac{q^{(n)}}{p} - s^{(n)})$ cannot converge, proving the second part of the claim. $\square$

### B.2. Discretisation approach

We define the notion of a *discretiser*, a measurable function $k\colon \mathbb{R}^D \to A$ where $A$ is a finite set the members of which will be called *cells*. We will consider discretisers that divide each axis of $\mathbb{R}^D$ into two half-intervals in the tails and many equal sized intervals in the middle; the size of these will be denoted by $\Delta$. Thus if $k$ divides a single axis into $M$ cells, the total number of cells in $\mathbb{R}^D$ will be $M^D$. We will consider sequences of discretisers $(k_n)_{n\in\mathbb{N}}$ where each $k_n$ produces discretisation which is a refinement of the previous one, i.e. it only divides existing cells into smaller ones.

We say that a sequence of discretisers is *asymptotically exact* if for every $x \in \mathbb{R}^D$ we have,

$$\bigcap_{n\in\mathbb{N}}\bigcap_{a\in A^{(n)}\,:\,k_n(x)=a}k_n^{-1}(a) = \{x\},$$

i.e. any two distinct points will end up in different cells eventually. With a slight abuse of notation, we abbreviate this as $\lim_{n\to\infty}k_n(x) = \{x\}$.

We further define a function $x_n\colon A^{(n)} \to \mathbb{R}^D$ which accepts a cell and returns an element that maps to that particular cell; such function must exist by the axiom of choice.

Finally, we denote the *quantised densities* w.r.t. the counting measure for P and Q respectively by $p^{(n)}(a) = P(k_n^{-1}(a))$ and $q^{(n)}(a) = Q(k_n^{-1}(a))$.

**Proposition 21.** *Consider an asymptotically exact sequence of discretisers* $(k_n)_{n\in\mathbb{N}}$*, the corresponding sequence of finite spaces* $(A^{(n)})_{n\in\mathbb{N}}$*, and discretisation intervals* $(\Delta_n)_{n\in\mathbb{N}}$*. Let S be at most countable and all the relevant aforementioned assumptions hold. We will consider two cases:* $\log p \in L^1(Q)$ *and* $\log p \notin L^1(Q)$.

*Then,*

$$\lim_{n\to\infty}\left\{\mathrm{KL}\left(Q^{(n)}\|P^{(n)}\right) - s^{(n)}\right\} = \mathbb{E}_{\widetilde{Q}}\left(\log\frac{q}{p|_S}\right),$$

*with* $s^{(n)} = -D\log(\Delta_n)$.

*Proof of Proposition 21.* By assumption, $\mathrm{diam}(S) < \infty$ and thus we can find a compact set $K \subset \mathbb{R}^D$ s.t. $S \subset K$. W.l.o.g. define $R_+ \supset K$ to be the smallest hyper-rectangle of strictly positive Lebesgue measure s.t. it can be padded out by hypercubes with side $\Delta_1$ (by extending the lengths of sides of $R$ to be positive multiples of $\Delta_1$; by the assumption that each $k_n$ refines existing cells, and that the cells are equal sized, $k_n(R_+)$ will only produce equal sized cells for all $n \in \mathbb{N}$). $R_+$ exists by the Heine–Borel theorem.

The $n^{\text{th}}$ discretised KL is defined as,

$$\mathrm{KL}\left(\mathrm{Q}^{(n)} \parallel \mathrm{P}^{(n)}\right) = \sum_{a \in A^{(n)}} q^{(n)}(a) \log \frac{q^{(n)}(a)}{p^{(n)}(a)}.$$

From now on, we will drop the input to the individual quantised densities unless confusion may arise.

We start with the case $\log p \in \mathrm{L}^1(\mathrm{Q})$. By Lemma 22, $\log p|_S \in \mathrm{L}^1(\widetilde{\mathrm{Q}})$. Because we assumed that $\log q \in \mathrm{L}^1(\widetilde{\mathrm{Q}})$,

$$\mathbb{E}_{\widetilde{\mathrm{Q}}} \log \tfrac{q}{p|_S} = \mathbb{E}_{\widetilde{\mathrm{Q}}}(\log q) - \mathbb{E}_{\widetilde{\mathrm{Q}}}(\log p|_S),$$

by Theorem 4.1.10 in (Dudley, 2002), and thus we can focus on the negative entropy and cross-entropy terms separately.

Starting with the negative entropy term, notice that for any $x \in S$, we have $q^{(n)}(k_n(x)) \to \widetilde{\mathrm{Q}}(\{x\})$, as for any $x' \in S \setminus \{x\}$, $\widetilde{\mathrm{Q}}(\{x'\}) > 0$ and there exists $\mathrm{N} \in \mathbb{N}$ s.t. $\|x - x'\|_2 > \sqrt{\mathrm{D}}\Delta_n$ (the maximum distance of points in a single cell) for all $n \geq \mathrm{N}$. Thus $q^{(n)}(k_n(x)) \downarrow \widetilde{\mathrm{Q}}(\{x\})$ by being a monotonically decreasing sequence with the least upper bound equal exactly to $\widetilde{\mathrm{Q}}(\{x\})$. Note that by assumption $\widetilde{\mathrm{Q}}(\{x\}) = q(x)$ where $q$ is the density $\widetilde{\mathrm{Q}}$ of w.r.t. the counting measure on $\mathbb{Q}^D$, and thus $q^{(n)}(k_n(x)) \downarrow q(x)$.

The following insight will help us:

$$\sum_{a \in A^{(n)}} q^{(n)}(a) h(a) = \int q(x) h(k_n(x)) m_S(dx), \quad (14)$$

for any $h \colon A^{(n)} \to \mathbb{R}$; note that the definition of $A^{(n)}$ makes $h(k_n(x))$ a simple function and thus measurable which means the r.h.s. is well-defined. We can thus use continuity and monotonicity of the logarithm to establish $\log q^{(n)}(k_n(x)) \downarrow \log q(x)$ pointwise and the fact that $\log q^{(n)}(k_n(x)) \leq 0$ as $q^{(n)}(k_n(x)) \leq 1, \forall x$, and apply the monotone convergence theorem to establish,

$$\sum_{A^{(n)}} q^{(n)} \log q^{(n)} \downarrow \int q \log q \, dm_S.$$

We now turn to the cross-entropy term. Because $R_+$ is compact, we can define,

$$\alpha_n := \max_{a \in k_n(R_+)} \left| \sup[\log p(k_n^{-1}(a))] - \inf[\log p(k_n^{-1}(a))] \right|,$$

and observe $\alpha_n \downarrow 0$ as $n \to \infty$ because $\log p$ is continuous, and thus uniformly continuous on $R_+$. Notice,

$$\left| \sum_{a \in A^{(n)}} q^{(n)}(a)(\log[p^{(n)}(a)] - \log[p(x_n(a))\Delta_n^D]) \right|$$
$$\leq \sum_{a \in A^{(n)}} q^{(n)}(a) \left| \log[p^{(n)}(a)] - \log[p(x_n(a))\Delta_n^D] \right|$$
$$\leq \sum_{a \in A^{(n)}} q^{(n)}(a) \alpha_n \leq \alpha_n,$$

using that $q^{(n)} = 0$ outside of $k_n(R_+)$. Because $\alpha_n \downarrow 0$ as $n \to \infty$, we can approximate $\log[p^{(n)}(a)\Delta_n^D]$ by $\log p(x_n(a)) + \mathrm{D} \log \Delta_n$.

Since $\lim_{n \to \infty} k_n(x) = \{x\}$ by assumption, we have $x_n(k_n(x)) \to x$ pointwise by $\|x - x'\|_2 \leq \sqrt{\mathrm{D}}\Delta_n$ for any $x'$ s.t. $k_n(x) = k_n(x')$. By continuity of the logarithm, $\log p(x_n(k_n(x))) \to \log p(x)$ pointwise (i.e. $\log p(x_n(a))$ can be substituted for the function $h(a)$ in Equation (14)). Because $R_+$ is compact, we can define $\kappa := \sup_{R_+} |\log p|$ which will be finite by the continuity of $\log p$. Hence $|\log p(x_n(k_n(x)))| \leq \kappa$, and we can apply the dominated convergence theorem:

$$\sum_{a \in A^{(n)}} q^{(n)}(a) \log p(x_n(a)) \to \int q \log p|_S \, dm_S.$$

Putting the results in previous paragraphs together, we arrive at the following limit,

$$\sum_{A^{(n)}} q^{(n)} \log \frac{q^{(n)}}{p^{(n)}} + \mathrm{D} \log \Delta_n \to \int q \log \frac{q}{p|_S} \, dm_S,$$

where we are implicitly using the previously derived equality $\mathbb{E}_{\widetilde{\mathrm{Q}}} \log \frac{q}{p|_S} = \mathbb{E}_{\widetilde{\mathrm{Q}}}(\log \mathrm{Q}) - \mathbb{E}_{\widetilde{\mathrm{Q}}}(\log p|_S)$.

It remains to investigate the case $\log p \notin \mathrm{L}^1(\mathrm{Q})$. Notice that our proof of convergence of $(\mathbb{E}_{\mathrm{Q}^{(n)}} \log p^{(n)} + \mathrm{D} \log \Delta_n)$ to $\mathbb{E}_{\widetilde{\mathrm{Q}}} \log p|_S$ is independent of $\log p \in \mathrm{L}^1(\mathrm{Q})$ and is facilitated using the dominated convergence theorem. The dominated convergence theorem states that the pointwise limit itself must be integrable, and thus the case $\log p \notin \mathrm{L}^1(\mathrm{Q})$ is never realised under our assumptions by Lemma 22. □

### B.3. Shared auxiliary lemmas

**Lemma 22.** *For* $\mathrm{Q}$ *a probability measure on* $(\mathbb{R}^D, \mathcal{B})$, $\widetilde{\mathrm{Q}}$ *its restriction to* $(S, \mathcal{B}_S)$, *and a Borel measurable function* $f \colon \mathbb{R}^D \to \mathbb{R}$, *the following holds,*

$$\mathbb{E}_{\mathrm{Q}} f = \mathbb{E}_{\widetilde{\mathrm{Q}}} f|_S,$$

*with* $f|_S$ *being the restriction of* $f$ *to* $S$.

*Proof of Lemma 22.* Because $\mathbb{E}_Q f = \int f \, dQ = \int f^+ \, dQ - \int f^- \, dQ$, with $f^+ = f \vee 0$ and $f^- = -(f \wedge 0)$, by definition of the Lebesgue integral, we can w.l.o.g. assume $f \geq 0$ so that $\int f \, dQ = \sup\{\int g \, dQ \colon 0 \leq g \leq f, g \text{ simple}\}$. For any simple $g$, using $\mathrm{supp}(Q) = S$,

$$\int_{\mathbb{R}^D} g \, dQ = \int_{\mathbb{R}^D} \sum_{j=1}^n a_j \mathbb{I}_{B_j} \, dQ = \sum_{j=1}^n a_j Q(B_j)$$

$$= \sum_{j=1}^n a_j Q(B_j \cap S) = \sum_{j=1}^n a_j \widetilde{Q}(B_j \cap S)$$

$$= \int_S \sum_{j=1}^n a_j \mathbb{I}_{B_j} \, d\widetilde{Q} = \int_S g|_S \, d\widetilde{Q},$$

with $\{a_j\}_{j=1}^n \subset \mathbb{R}, n \in \mathbb{N}$. Taking the supremum on both sides establishes $\int_{\mathbb{R}^D} f \, dQ = \int_S f|_S \, d\widetilde{Q}$. $\square$

## C. Proofs for Section 5

*Proof of Proposition 6.* Let us first check the assumptions of Proposition 9. Clearly, the respective densities are continuous and bounded. Furthermore, the entropy of Q is equal to $\frac{1}{2} \log \det_*(2\pi e AVA^T)$ where $\det_*$ is the pseudo-determinant, and thus $\log q \in L^1(\widetilde{Q})$. It is also clear that $\mathbb{E}_{Z \sim Q} \|Z\|_2^2 < \infty$ by the relation of the squared norm of Gaussian random variables and the $\chi^2$ distribution. We will use Proposition 10 to ensure that the collection of random variables $\{\log p(Z^{(n)})\}$, $Z^{(n)} \sim Q^{(n)}$, is uniformly integrable. Observe that for all $z \in \mathbb{R}^D$,

$$|\log p(z)| \leq c + \frac{1}{2}|z^T \Sigma^{-1} z| \leq c + \frac{\|z\|_2^2}{2\gamma_0},$$

where $c \in \mathbb{R}_+$ is a constant, and $\gamma_0$ is the lowest eigenvalue of $\Sigma$ which is higher than zero because $\Sigma$ is a (strictly) positive definite matrix by assumption. As we have already established $\mathbb{E}_{Z \sim Q} \|Z\|_2^2 < \infty$, Proposition 10 holds and thus Proposition 9 can be applied.

For fixed $A$, the Q distribution has support over the subspace $S = \{x \in \mathbb{R}^D \colon x = Az, z \in \mathbb{R}^K\}$. If $z \sim \mathcal{N}_K(0, V)$, then $Az \sim \mathcal{N}_D(0, AVA^T)$. Hence we can perform substitution which reduces QKL to,

$$\int_{\mathbb{R}^K} \phi_{0,V}(z) \log \frac{\phi_{0,V}(z)}{\phi_{0,A^T \Sigma A}(z)} \lambda^K(dz)$$

where we have used the identity $(A^T \Sigma^{-1} A)^{-1} z = A^T \Sigma A z$ for any $z \in \mathbb{R}^K$. The first term equals $-\frac{1}{2} \log |V| = -\frac{1}{2} \sum_{k=1}^K \log V_{kk}$ up to an additive constant, and the second to $\mathrm{Tr}(A^T \Sigma^{-1} AV)$ up to another additive constant. For a constant $c \in \mathbb{R}$, the integral equals,

$$c - \frac{1}{2} \sum_{k=1}^K \log V_{kk} + \frac{1}{2} \mathrm{Tr}(A^T \Sigma^{-1} AV).$$

The second term can be rewritten as,

$$\mathrm{Tr}(A^T \Sigma^{-1} AV) = \sum_{k=1}^K V_{kk} a_k^T \Sigma^{-1} a_k,$$

where $a_k$ is the $k^{\text{th}}$ column of the $A$ matrix. Because this is an additive loss term in the above QKL, and $V_{kk} > 0$ by the construction of $S$, it is minimised when the $a_k$ vectors are aligned with the top K eigenvectors of $\Sigma$ because then $a_k^T \Sigma^{-1} a_k = 1/\gamma_k$ which will be lowest for the highest eigenvalues $\gamma_k$ of $\Sigma$. Differentiating the objective w.r.t. $V_{kk}$ after substituting the optimal $A$ yields,

$$-\frac{1}{2} \frac{1}{V_{kk}} + \frac{1}{2} \frac{1}{\gamma_k}.$$

Setting to zero, we see that $V_{kk} = \gamma_k$, i.e. matching the eigenvalues of $\Sigma$ is the optimal solution. $\square$

*Proof of Proposition 6.* The $n^{th}$ KL is up to an additive constant equal to,

$$\mathcal{L} := \mathrm{Tr}\left((AVA^T + \tau^{(n)} I)\Sigma^{-1}\right) - \log\left|AVA^T + \tau^{(n)} I\right|.$$

Using some matrix calculus identities from (Petersen et al., 2008), the derivatives w.r.t. the individual parameters are,

$$\nabla_A \mathcal{L} = \Sigma^{-1} A - (AVA^T + \tau^{(n)} I)^{-1} A,$$

$$\nabla_{\mathrm{diag}(V)} \mathcal{L} = \mathrm{diag}[A^T(\Sigma^{-1} - (AVA^T + \tau^{(n)} I)^{-1})A].$$

Defining a new diagonal matrix $\widehat{V}_{kk}^{(n)} = V_{kk} + \tau^{(n)}$, and using the orthogonality of $A$'s columns, we have,

$$\nabla_A \mathcal{L} = \Sigma^{-1} A - A(\widehat{V}^{(n)})^{-1},$$

$$\nabla_{\mathrm{diag}(V)} \mathcal{L} = \mathrm{diag}[A^T \Sigma^{-1} A - (\widehat{V}^{(n)})^{-1}].$$

Setting the first formula above to zero leads to an eigenvector problem, hence we know that the columns of $A$ must be eigenvectors of $\Sigma$. Setting the second formula to zero yields,

$$V_{kk} = (a_k^T \Sigma^{-1} a_k)^{-1} - \tau^{(n)}.$$

which after substitution of $a_k$ by an eigenvector leads to $V_{kk} = \gamma_k - \tau^{(n)}$ where $\gamma_k$ is the eigenvalue for the $k^{th}$ substituted eigenvector. By substituting into $\mathcal{L}$,

$$c + \sum_{k=1}^K \frac{\gamma_k}{\gamma_k} - \log(\gamma_k - \tau^{(n)}),$$

where $c$ is a constant, we see that to the objective is minimised when the eigenvectors corresponding to the highest eigenvalues are selected. Hence the solution for $A$ is the same as for PCA for all $n \in \mathbb{N}$, and $|\gamma_k - (\gamma_k - \tau^{(n)})| \to 0$ as $n \to \infty$. The optimal solution thus converges to the PCA/QKL in Frobenius/Euclidean distance. $\square$